\def\year{2022}\relax
\documentclass[letterpaper]{article} 
\usepackage{aaai22}  
\usepackage{times}  
\usepackage{helvet}  
\usepackage{courier}  
\usepackage[hyphens]{url}  
\usepackage{graphicx} 

\usepackage{stmaryrd}
\usepackage{amsmath}
\usepackage[noend]{algorithmic}
\usepackage{algorithm}
\usepackage{booktabs}
\usepackage{multirow}
\usepackage{wrapfig}
\usepackage{bbm}
\usepackage{siunitx}
\newcolumntype{P}[1]{>{\centering\arraybackslash}p{#1}}
\usepackage{natbib}  
\usepackage{caption} 
\DeclareCaptionStyle{ruled}{labelfont=normalfont,labelsep=colon,strut=off} 
\usepackage{algorithm}
\usepackage{algorithmic}

\usepackage{newfloat}
\usepackage{listings}

\title{What can we Learn Even From the Weakest? \\ Learning Sketches for Programmatic Strategies}
\author{
Leandro C. Medeiros,\textsuperscript{\rm 1}\equalcontrib 
David S. Aleixo,\textsuperscript{\rm 1}\equalcontrib 
Levi H. S. Lelis\textsuperscript{\rm 2} \\
}
\affiliations{
\textsuperscript{\rm 1}Departamento de Inform\'atica, Universidade Federal de Vi\c{c}osa, Brazil\\
\textsuperscript{\rm 2}Department of Computing Science, Alberta Machine Intelligence Institute (Amii), University of Alberta, Canada \\
}

\usepackage{bibentry}

\begin{document}

\newcommand{\sygus}[0]{SyGuS}
\newcommand{\mc}{\mathcal}
\newcommand*\GOR{\ |\ }

\maketitle

\begin{abstract}
In this paper we show that behavioral cloning can be used to learn effective sketches of programmatic strategies. We show that even the sketches learned by cloning the behavior of weak players can help the synthesis of programmatic strategies. This is because even weak players can provide helpful information, e.g., that a player must choose an action in their turn of the game. If behavioral cloning is not employed, the synthesizer needs to learn even the most basic information by playing the game, which can be computationally expensive. We demonstrate empirically the advantages of our sketch-learning approach with simulated annealing and UCT synthesizers. We evaluate our synthesizers in the games of Can't Stop and MicroRTS. The sketch-based synthesizers are able to learn stronger programmatic strategies than their original counterparts. Our synthesizers generate strategies of Can't Stop that defeat a traditional programmatic strategy for the game. They also synthesize strategies that defeat the best performing method from the latest MicroRTS competition.  
\end{abstract}

\section{Introduction}

One needs to search in large program spaces to synthesize effective programmatic strategies. 
In addition to dealing with large spaces, synthesizers often lack effective functions for guiding the search. This is in contrast with neural methods, where gradient information is available to guide the search. The problem of neural methods is their lack of interpretability. Despite being elusive, we can often understand, verify, and even manually modify programmatic strategies. 


In this paper we show that behavioral cloning~\cite{Bain96aframework} can be used to learn program sketches~\cite{solar2009sketching} to speed up the synthesis of strong programmatic strategies. Sketches are incomplete programs that serve as starting points for synthesis. 
%
%
We investigate the use of this sketch learning approach with synthesizers employing Simulated Annealing (SA)~\cite{simulated_annealing} and UCT~\cite{Kocsis06banditbased} as search algorithms and evaluate them in the context of computing a best response for a target strategy in two-player zero-sum games. Specifically, we evaluate our methods in the board game of Can't Stop and in the real-time strategy (RTS) game of MicroRTS. We show that our methods can be effective even when cloning the behavior of weak players, such as 
a player that chooses their actions at random in the game of Can't Stop. 
Our sketch learning method can be effective when the cloned strategy is weak because even such strategies might convey information 
that is helpful for the synthesis of programmatic strategies, such as the program structure required to decide when to stop playing in Can't Stop or when to build a specific structure in MicroRTS. 

We evaluated our sketch-based SA and UCT methods by synthesizing approximated best responses to a known programmatic strategy for Can't Stop~\cite{Glenn2009AGH} and to COAC, the winner of the latest MicroRTS Competition~\cite{MicroRTSCompetition2020}, on four maps. Our sketch-based SA synthesized strong strategies in all settings tested. As a highlight, it synthesized a strategy that defeats COAC on large maps of MicroRTS; none of the strategies synthesized by the baselines were able to defeat COAC on large maps.



\section{Related Works}

Our work is related to methods for synthesizing programmatic policies, 
in particular those that use some form of behavioral cloning such as imitation learning~\cite{Schaal1999IsIL}. \citeauthor{BastaniPS18}~\shortcite{BastaniPS18} presented an algorithm that uses a variant of the imitation learning algorithm DAgger~\cite{pmlr-v15-ross11a} to distill a high-performing neural policy into an interpretable decision tree. 

\citeauthor{VermaMSKC18}~\shortcite{VermaMSKC18} use DAgger and a neural policy to help with the synthesis of programmatic policies. The actions the neural policy chooses on a set of states are used in a Bayesian optimization procedure for finding suitable constant values for the programmatic policies. 
\citeauthor{VermaProjection2019}~\shortcite{VermaProjection2019} use a similar approach, but the neural model is trained so that it is not ``too different'' from the synthesized policies, with the goal of easing the optimization task. 

We differ from previous work in that we do not assume the oracle is available for queries as in DAgger-like methods. We also do not assume that the oracle is a neural network as \citeauthor{VermaProjection2019}~\shortcite{VermaProjection2019} do. 
For example, in our experiments we use a data set from a human oracle. We also do not require the strategy to be cloned to be high performing; we are able to learn effective sketches even from weak strategies. 

\citeauthor{Marino2021}~\shortcite{Marino2021} introduced Lasi, a method that uses behavioral cloning to simplify the language used for synthesis. Lasi removes from the language the instructions that are not needed to clone a strategy. Lasi can be used with our sketch-learning methods by simplifying the language to only then learn a sketch. Lasi is not as general as our methods because it cannot be applied to domains in which all symbols in the language are needed, 
such as Can't Stop. 

Others have synthesized programs to serve as evaluation functions~\cite{Benbassat11},  
but not to serve as complete strategies. Others explored the synthesis of strategies for cooperative games~\cite{canaan2018evolving} and single-agent problems~\cite{butler2017synthesizing,de2018evolving}. These methods can potentially benefit from our sketch-learning methods. 

While most previous work assume that the user provides the program sketch~\cite{solar2009sketching}, 
\citeauthor{pmlr-v97-nye19a}~\shortcite{pmlr-v97-nye19a} use a neural model to generate sketches. 
Their approach is designed to solve program synthesis tasks, where one synthesizes a program mapping a set of input values to the desired output values; we synthesize strategies.  
Also, we are unable to train a neural model for sketch generation because the amount of data we consider is insufficient for training (we use data sets with state-actions of as few as 3 matches).

\section{Problem Definition}

Let $G$ be a sequential two-player zero-sum game defined by a set $S$ of states, a pair of players $P = \{i, -i\}$, an initial state $s_{\text{init}}$ in $S$, a function $A_i(s)$ that receives a state $s$ and returns the set of actions player $i$ can perform at $s$, and a function $U_i(s)$ that returns the utility of player $i$ at $s$. Since $G$ is zero sum, $U_i(s) = -U_{-i}(s)$. 
A strategy for player $i$ is a function $\sigma_i : S \rightarrow A_i$ mapping a state $s$ to an action $a$. 
A programmatic strategy is a computer program encoding a strategy $\sigma$. 
The value of the game for state $s$ is denoted by $U(s, \sigma_i, \sigma_{-i})$, which returns the utility of player $i$ if $i$ and $-i$ follow the strategies given by $\sigma_i$ and $\sigma_{-i}$. 
We also call a match a game played between two strategies. 

We consider programmatic strategies written in a domain-specific language (DSL)~\cite{van2000domain}. Let $D$ be a DSL and $\llbracket D \rrbracket$ be the set of programs written in $D$. 
The best response for a strategy $\sigma_{-i}$ in $\llbracket D \rrbracket$ is a strategy that maximizes player $i$'s utility against $\sigma_{-i}$, i.e., $\max_{\sigma_{i} \in \llbracket D \rrbracket} U(s_{init}, \sigma_{i}, \sigma_{-i})$. The computation of a best response for a fixed strategy is a basic operation in game theory approaches such as iterated best response for approximating a Nash equilibrium profile~\cite{lanctotunifying}. 

In this paper we evaluate different search methods for synthesizing a best response to a strategy. We provide a game $G$, a DSL $D$ and a strategy $\sigma_{-i}$ and the synthesizer searches in the space defined by $D$ and returns an approximated best response $\sigma_i$ in $\llbracket D \rrbracket$ to $\sigma_{-i}$. We also consider the setting in which a data set with state-action pairs $L = \{(s_j, a_j)\}_{j=1}^m$ with the actions $a_j$ a player takes at states $s_j$ is available. 

\section{Synthesis of Programmatic Strategies}

In this section we review DSLs and explain how SA and UCT can be used to synthesize programmatic strategies. While SA and UCT have been applied to program synthesis tasks, e.g., \cite{programsynthesis_sa,cazenave2013}, this is the first time these approaches are applied to synthesize programmatic strategies, so we describe them in detail. 

\subsection{Domain-Specific Languages}

\begin{figure}[h]
\begin{minipage}{0.24\textwidth}
\begin{align*}
I &\to \, C \GOR \text{if}(B) \text{ then } C \, \\
C &\to c_1 \GOR c_2 \\
B &\to b_1 \GOR b_2 
\end{align*}
\end{minipage}
\begin{minipage}{0.21\textwidth}
\includegraphics[width=\linewidth,keepaspectratio=true]{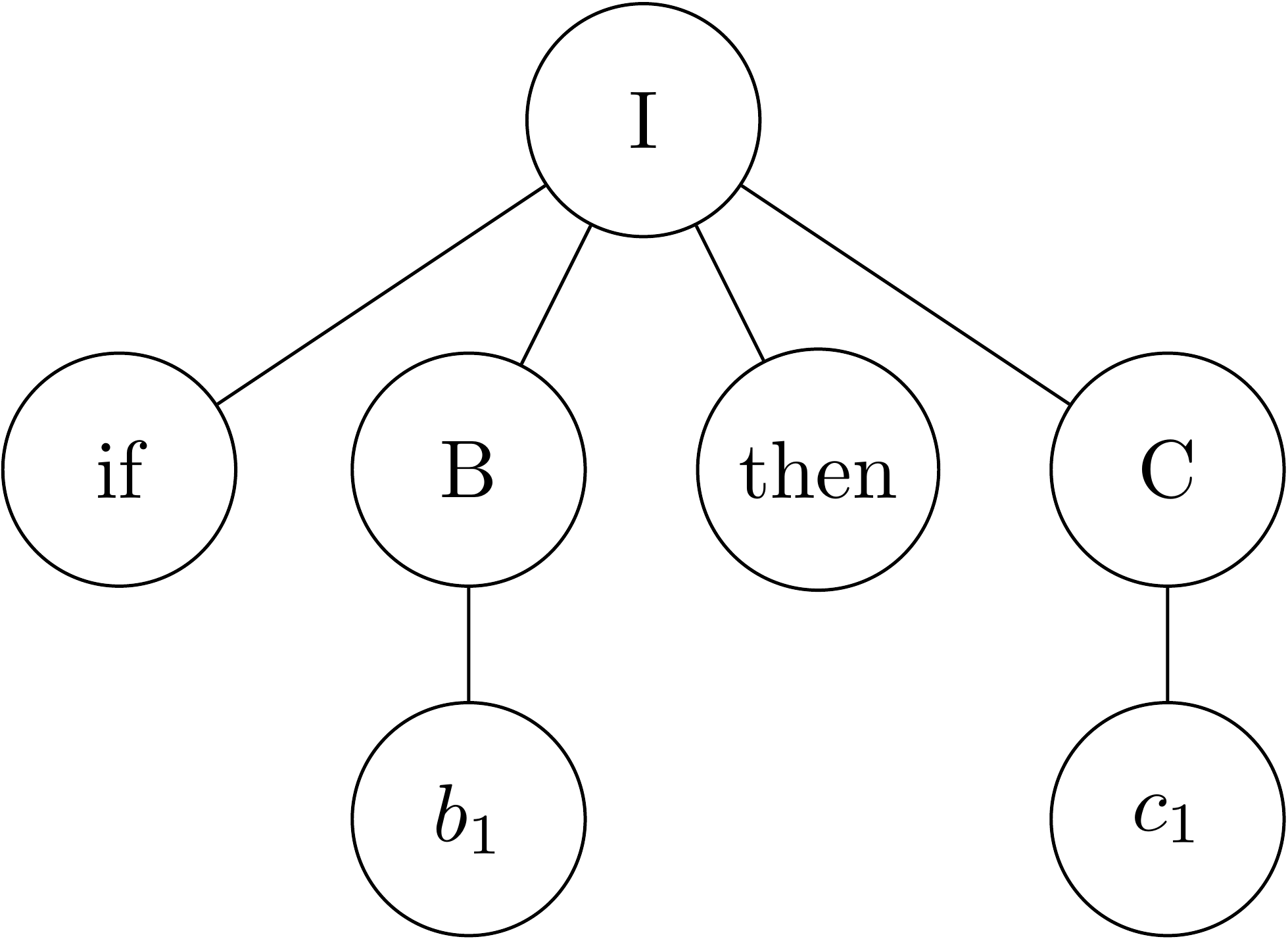}
\end{minipage}
\caption{DSL (left) and AST for ``if $b_1$ then $c_1$'' (right).}
\label{fig:dsl}
\end{figure}

A DSL is defined as a context-free grammar $(V, \Sigma, R, I)$, where $V$, $\Sigma$, and $R$ are sets of non-terminals, terminals, relations defining the 
production rules of the grammar, respectively. $I$ is the grammar's start symbol. Figure~\ref{fig:dsl} shows a DSL where 
$V = \{I, C, B\}$, $\Sigma = \{c_1, c_2, b_1, b_2$ if, then$\}$, $R$ are the relations (e.g., $C \to c_1$), and $I$ is the start symbol. 

The DSL allows programs with a single command ($c_1$ or $c_2$) and programs with branching. We represent programs as abstract syntax trees (AST), where the root of the tree is $I$, the internal nodes are non-terminals and leaf nodes are terminals.
Figure~\ref{fig:dsl} shows an example of an AST, where leaves are terminal symbols and internal nodes are non-terminals.  

\subsection{Simulated Annealing for Synthesis of Strategies}

SA is a local search algorithm that uses a temperature parameter to control the greediness of the search. SA behaves like a random walk in the beginning of the search and more like hill climbing later in search. We use SA to approximate a programmatic best response to a target strategy $\sigma_{-i}$, i.e., SA approximates a solution to $\arg\max_{\sigma_{i} \in \llbracket D \rrbracket} U(s_{init}, \sigma_{i}, \sigma_{-i})$. 

SA starts with a program that is randomly generated as follows. We start with $I$ and we replace it with a randomly chosen production rule for $I$; we then repeatedly replace a non-terminal symbol in the generated program with a random and valid production rule; we stop when the program contains only terminals.  
For example, the production rules used to obtain program ``if $b_1$ then $c_1$'' are: $I \rightarrow \text{if}(B) \text{ then } C \text{ else } C$; $B \rightarrow b_1$; $C \rightarrow c_1$. 

Once the initial program $p$ is defined, SA generates a neighbor $p'$ of $p$ by changing a subtree in $p$'s AST. We randomly choose a non-terminal symbol $n$ in the AST (all non-terminal symbols can be chosen with equal probability) and we replace the subtree rooted at $n$ with a subtree that is generated with the same procedure used to generate the initial program. For example, if the subtree rooted at $C$ (Figure~\ref{fig:dsl}) is chosen and we replace it with $c_2$,  
then $p'$ is ``if $b_1$ then $c_2$.'' SA decides if it accepts or rejects $p'$. If it accepts, then $p'$ is assigned to $p$ and the process is repeated. If it rejects, SA repeats the procedure by generating another neighbor of $p$. The probability in which SA accepts $p'$ is given by 
\begin{equation*}
\min\bigg(1, \exp\bigg(\frac{\beta \cdot (\Psi(p') - \Psi(p))}{T_j}\bigg)\bigg) \,.
\end{equation*}
Here, $T_j$ is the temperature at iteration $j$, and $\Psi$ is an evaluation function. In program synthesis tasks, $\Psi(p)$ counts the number of input examples that $p$ correctly maps to the desired output~\cite{AlurBJMRSSSTU13}. In the context of games, $\Psi(p)$ returns the utility of $p$ against the opponent $\sigma_{-i}$. If $\Psi(p') \geq \Psi(p)$, then SA accepts $p'$ with probability $1.0$. Otherwise, the probability of acceptance depends on $T_j$ and $\beta$. $\beta$ is an input parameter that allows us to adjust how greedy SA is; larger values of $\beta$ result in a greedier search by more often rejecting programs with small $\Psi$-values. Larger values of $T_j$ make the search less greedy since large $T$-values increase the chances of accepting $p'$. The initial temperature, $T_1$, is an input parameter and $T_j$ is computed according to the schedule 
$T_j = \frac{T_1}{ (1 + \alpha \cdot j)}$. 
Once the temperature becomes smaller than $\epsilon$, we stop searching and the program with largest $\Psi$-value encountered in search is returned as the SA's approximated best response to $\sigma_{-i}$. In our experiments we run SA multiple times, while we have not exhausted the time allowed for synthesis, and we initialize the search with the program returned in the latest run as it often allows the search to start in a more promising region of the space. 

\subsection{UCT for Synthesis of Strategies}

UCT grows a search tree while exploring the space. Each node in the tree represents a program, which can be complete or incomplete. A program is complete if all leaves in its AST are terminals. The root of the UCT tree represents the incomplete program of the initial symbol $I$. The children of a node $n$ in the UCT tree are the programs that can be generated by applying a production rule to the leftmost non-terminal symbol of the program $n$ represents. For example, if $n$ represents  ``$\text{if}(B) \text{ then } C$'', then its children represent ``$\text{if}(b_1) \text{ then } C$'' and ``$\text{if}(b_2) \text{ then } C$'' because $B$ is the leftmost non-terminal symbol of the program $n$ represents. 

UCT operates in four steps: selection, expansion, simulation, and backpropagation. The selection step starts at the root of the tree and it chooses the $j$-th child that maximizes 
    $\bar{X}_j + K \sqrt{\frac{\log(N)}{N_j}}$. 
Here, $\bar{X}_j$ is the average evaluation value of the $j$-th child of $n$, $N$ is the number of times node $n$ was visited in previous selection steps,  $N_j$ is the number of times $n$ was visited and the $j$-th child was selected, and $K$ is an exploration constant. The first term of the equation is an exploitation term as it favors the child with highest average evaluation value; the second term is an exploration term. 

The selection step stops when it encounters a node $n$ with at least one child $n'$ that is not in the UCT tree. In the expansion step, UCT adds $n'$ to the tree; if more than one child is not in the UCT tree, the algorithm chooses one arbitrarily. The simulation step applies a policy to turn $n'$ into a complete program, which is then evaluated with $\Psi$. Finally, the backpropagation step updates the $X_j$-values of all nodes visited in the selection step with the $\Psi$-value from the simulation step. The four steps are repeated multiple times and UCT returns the program with largest $\Psi$-value, among all programs evaluated during search, when it reaches a user-specified time limit. UCT caches the $\Psi$-values of programs that were evaluated in previous iterations of the algorithm. 
The UCT tree might grow to include complete programs (i.e., nodes with no children). If the selection step ends at a complete program, it performs no expansion and returns the cached $\Psi$-value of $n$ in the backpropagation step. 

We use a single run of SA as UCT's simulation policy. When running SA as simulation policy for an incomplete program $p$, the neighbors of a program can only be obtained by changing the subtrees of the AST that are rooted at a non-terminal leaf node. For example, if $p$ is ``$\text{if}(B) \text{ then } C$'', then the neighbors of $p$ can be obtained by changing only $B$ and $C$, but not the root of the AST, $I$, because $I$ is not a leaf in the AST. This constraint ensures that the simulation policy does not change the structure of $p$, which is defined by the production rules along the path in the UCT tree.

\section{Learning Sketches with Behavioral Cloning}

We consider the setting in which the synthesizer receives as input a data set of state-action pairs $L = \{(s_j, a_j)\}_{j=1}^m$ with actions chosen by strategy $\sigma_o$ for states of one or more matches of the game. We use this data set to learn a sketch to speed up the synthesis of strong programmatic strategies. 

SA and UCT can be used to clone the behavior of $\sigma_o$ 
by replacing $\Psi$ with an evaluation function $C(L, p)$ that receives the data set $L$ and a program $p$ and returns a score of how well $p$ clones $\sigma_o$. Note, however, that cloning the behavior of $\sigma_o$ can result in weak strategies. 
This is because $\sigma_o$ might not be represented in $\llbracket D \rrbracket$. Or the data set $L$ is limited and one needs to perform DAgger-like queries~\cite{pmlr-v15-ross11a} to augment it and $\sigma_o$ might not be available for such queries (e.g., $\sigma_o$ is a human player who is unavailable). Or $\sigma_o$ is a weak strategy and exactly cloning its behavior would result in a weak strategy. Instead of learning a strategy directly with behavioral cloning, we use it to learn a sketch that helps the synthesis process of a strong strategy. 

Sketch-learning methods can be more effective than those that optimize for $\Psi$ directly for two reasons. First, $\Psi$ can be computationally more expensive than $C$. Using $C$ to learn parts of the programmatic strategy  will tend to be more efficient than to learn the entire strategy with $\Psi$. Second, the function $C$ can offer a denser signal for search (e.g., the neighbor $p'$ of $p$ might not defeat $\sigma_{-i}$, but it might have a higher $C$-score, which can be helpful to guide the search).

\subsection{Sketch Learning with UCT}

We run UCT with the evaluation function $C(L, p)$ for a number of iterations and, whenever we find a complete program $p$ with a $C$-value larger than the current best solution, we evaluate it with $\Psi$. We call this search the sketch-search. Once we reach a time limit, we use the program found in the sketch-search with the largest $\Psi$-value to initialize a second UCT search, which we call 
best response (BR)-search. 

Let $p$ be the program with largest $\Psi$-value encountered in the sketch-search. The program is defined by a sequence of production rules that replace the leftmost non-terminal symbol in the sequence of partial programs, starting with the initial symbol of the DSL. 
We start the UCT tree of the BR-search with a branch that represents the production rules of $p$. For example, let ``$\text{if}(b_1) \text{ then } c_1$'' be the program $p$ with largest $\Psi$-value from the sketch-search. 
The UCT tree of the BR-search is initialized with the branch with nodes representing the programs: ``$I$'', ``$\text{if}(B) \text{ then } C$'', ``$\text{if}(b_1) \text{ then } C$'', and ``$\text{if}(b_1) \text{ then } c_1$''. We then perform a backpropagation step on the added branch with $\Psi(p)$, which was computed in the sketch-search. By adding the branch leading to $p$ to the tree of the BR-search we are biasing it to explore programs that share the structure of $p$. This is because the nodes along the added branch will likely have higher $\bar{X}$-values than other branches, specially in the first iterations of search. 

The branch added to the UCT tree of the BR-search acts as a sketch as defined in the literature~\cite{solar2009sketching} because it represents a program with ``holes'' that are filled by the BR-search. In our example, assuming that the $\Psi$-value of $p$ is somewhat large, the BR-search will be biased to explore the sketches that share the structure of $p$, such as ``$\text{if}(?) \text{ then } c_1$'' and ``$\text{if}(?) \text{ then } ?$'', where each question mark represents a hole that needs to be filled. 
Sketch learning provides a set of sketches with varied levels of detail (deeper nodes in the branch represent sketches with more information) that the BR-search explores while optimizing for $\Psi$. 

\subsection{Sketch Learning with Simulated Annealing}

Like with UCT, we run SA to clone $\sigma_o$ 
by using 
$C(L, p)$ as evaluation function. During search, every time we find a solution with better $C$-value, we also evaluate it with $\Psi$. Once we reach a time limit, SA returns the program $p$ with largest $\Psi$-value. We also call this search the sketch-search. We then use $p$ as the initial program of another SA search that optimizes for $\Psi$ directly, which we also refer to as the BR-search. We hypothesize that the program $p$ allows the BR-search to start in a more promising part of the space, because $p$ might have a structure that is similar to the structure of a program that approximates a best response to $\sigma_{-i}$.

While the branch added to the UCT tree of the BR-search can be seen as a set of sketches that are explored according to the prioritization defined by UCT, the connection between using program $p$ to initialize the BR-search and sketches is not as clear. We see the program $p$ as a \emph{soft sketch}, because it provides an initial structure to the synthesizer, but it does not explicitly specify a set of holes. Since SA can change any subtree of $p$'s AST, any subtree can be seen as a \emph{soft hole} of $p$. Some subtrees are more likely to be replaced than others due to SA's acceptance function, i.e., SA prefers to change subtrees that will result in an increase in $\Psi$-value. 


\subsection{Score Functions for Behavioral Cloning}

We use domain dependent functions $C(L, p)$ and describe them in the empirical section. We consider score functions that use both the state and actions in the data set $L = \{(s_j, a_j)\}_{j=1}^m$ and functions that use only the states in $L$, as in recent approaches on imitation learning from observations~\cite{torabi2018}. 

\section{Empirical Evaluation}

The goal of our evaluation is to verify if synthesizers that learn a sketch with behavioral cloning generate stronger approximated best responses to $\sigma_{-i}$ than their counterparts, that optimize directly for $\Psi$. 
All experiments were run on a single 2.4 GHz CPU with 8 GB of RAM and a time limit of 2 days.\footnote{The implementation of all algorithms used in our experiments is available at \url{https://github.com/leandrocouto/sketch-learning}.}
We use $\alpha = 0.9$, $\beta = 200$, $T_1 = 100$, $\epsilon = 1$, and $K=10$.

\subsection{Problem Domains}

We use the two-player versions of Can't Stop and MicroRTS. We chose these games because they have different features and allow for a diversity of scenarios. While Can't Stop is a stochastic game, MicroRTS is a deterministic game played with real-time constraints. The branching factor of Can't Stop is small (2 or 3 actions per state), while MicroRTS has an action space that can grow exponentially with the number of game components~\cite{lelis2020}. Finally, there exist strong human-written programmatic strategies for these games that we can use as $\sigma_{-i}$ in our experiments. 

\subsubsection{Can't Stop} The game of Can't Stop is played on a board with 11 columns, numbered from 2 to 12. The column 2 has 3 rows and the number of columns increases in size by 2 for every column until column 7, which has 13 rows. The number of rows decreases by 2, starting at column 8 until column 12, which also has 3 rows. The player who first conquers 3 columns wins the game. In each round of the game the player has 3 neutral tokens and they roll 4 six-sided dice. The player can place a neutral token in any column that is given by the combination of a pair of dice. A neutral token is then placed on the board, initially at the first row of the chosen column and later immediately above the player's permanent marker. The player can then decide to stop playing or to roll the dice again. If the player chooses the former, the neutral tokens are replaced by permanent tokens, thus securing that position on the board. If the player decides to roll the dice again, they are able to use the remaining neutral tokens, if there are any, or advance in columns in which they already have a neutral token placed. If the player does not have neutral tokens and the combination of dice only result in column numbers for which the player does not have a neutral token on, the player loses the neutral tokens and the other player starts their turn. A column is conquered when a player places a permanent token on the last row of a column. 

\citeauthor{Glenn2009AGH}~\shortcite{Glenn2009AGH} used a genetic algorithm to improve an existing programmatic strategy for Can't Stop~\cite{keller1986}. We use \citeauthor{Glenn2009AGH}'s strategy as $\sigma_{-i}$ and we call it GA. GA decides to stop playing in a turn of the game whenever the sum of scores of the neutral markers exceeds a threshold. GA defines a program for computing such a score (yes-no decision) and another program to decide in which column to advance next (column decision). 

We have developed a DSL for synthesizing programs for both the yes-no and column decisions. The DSL includes operators such as \texttt{map}, \texttt{sum}, \texttt{argmax}, and lambda functions. The DSL also includes a set of domain-specific functions such as a function for counting the number of rows a player has advanced in a turn. See our codebase for more information about the DSL.
The task is to synthesize a program that simultaneously solves the yes-no and the column decisions while maximizing the player's utility against GA. 

The $\Psi$ function for Can't Stop is the average number of victories of $p$ against $\sigma_{-i}$ on 1,000 matches. 
We run the sketch-search of SA for 1 hour and the BR-search for the remaining time. We run the sketch-search of UCT for 10 hours because UCT is slower than SA in exploring the space. We did not evaluate other time schedules for the synthesis of strategies with the sketch learning approaches. 

\subsubsection{MicroRTS} 
In MicroRTS each player controls a set of units of different types. 
Worker units can collect resources, build structures (Barracks and Bases), and attack opponent units. Barracks and Bases can neither attack opponents units nor move, but they can train combat units and Workers, respectively. Combat units can be of type Light, Heavy, or Ranged. These units differ in how long they survive a battle, how much damage they can inflict to opponent units, and how close they need to be from opponent units to attack them. We use a version of MicroRTS where the actions are deterministic and there is no hidden information. 
A match is played on a map and each map might require a different strategy for winning the game. 
We use four maps of different sizes, where the names in parenthesis are the names in the MicroRTS codebase:\footnote{\url{https://github.com/santiontanon/microrts}} 16$\times$16 (TwoBasesBarracks), 24$\times$24 (BasesWorkers), 32$\times$32 (BasesWorkers), and 64$\times$64 (BloodBath-B). The smallest and largest maps are from the 2020 MicroRTS Competition. 
We use the winner of the latest MicroRTS Competition, COAC, as $\sigma_{-i}$~\cite{MicroRTSCompetition2020}. COAC is a programmatic strategy written by humans. 

We have implemented a DSL similar to the one presented by \citeauthor{Marino2021}~\shortcite{Marino2021}. The DSL includes loops, conditionals, and a set of domain-specific functions that assign actions to units (e.g., build a barracks) and a set of  Boolean functions; see our codebase for details. 

The $\Psi$ function for MicroRTS is the average number of victories of $p$ against $\sigma_{-i}$ in 2 matches. Each map has two starting locations, so we run 2 matches alternating the players' starting location for fairness. MicroRTS does not require an explicit time schedule for splitting the time between the sketch-search and the BR-search. This is because both $\Psi$ and $C(L, p)$ are computed by having $p$ play 2 matches against $\sigma_{-i}$. The transition between sketch-search and BR-search occurs naturally if we define the evaluation function of the search algorithms as the $\Psi$ function with ties being broken according to $C(L, p)$. In the beginning of the synthesis the $\Psi$-value will be zero for all programs evaluated in search, but $C(L, p)$ quickly provides different values for different programs, which will guide the search toward helpful sketches. 

\subsection{Score Functions}

We consider an action-based score function where the score of $p$ is the fraction of actions that $p$ chooses at states in pairs $(s_j, a_j)$ of $L$ that match the action in the pair, i.e., $\sum_{(s_j, a_j) \in L} \mathbbm{1}[a_j = p(s_j)]/|L|$, where $\mathbbm{1}$ is the indicator function. We denote SA and UCT learning sketches with this score function as Sketch-SA(A) and Sketch-UCT(A). 

\subsubsection{Can't Stop} We use an observation-based score function that measures the percentage of permanent markers on the end-game state of a match that overlaps with the permanent markers obtained by a program $p$ on the match's end-game state if $p$ had played it. This score is computed by iterating through each state $s_j$ of a match in $L$ and applying the effects of actions $p(s_j)$ to an initially empty board of the game; once an end-game state $s_f$ is reached, we compute the percentage of overlapping permanent markers between $s_f$ and the end-game state in $L$. For example, if the player in the end-game state of a match in $L$ conquered columns 2, 3, and 7, and had one marker on column 12, and program $p$ conquered columns 2, 3, and had one marker on column 8, then the score is $(3 + 5)/(3 + 5 + 13 + 1 + 1) = 0.34$. Here, $(3 + 5)$ is the number of positions in the intersection of the end-game states and $(3 + 5 + 13 + 1 + 1)$ is the union of the positions. If $p$ and $\sigma_o$ return the same actions for all states in $L$, then the score is $1.0$. If $L$ has multiple matches, we return the average score across all matches. We denote SA and UCT using this score function as Sketch-SA(O) and Sketch-UCT(O). 

\subsubsection{MicroRTS} We use an observation-based score function that computes a normalized absolute difference between (i) the number of units and resources the strategy $p$ trains and collects in a match of $p$ against $\sigma_{-i}$ and (ii) the number of units and resources the strategy $\sigma_o$ trains and collects in a match in the data set $L$. Let $n_u$ and $n'_u$ be the number of units of type $u$ that $p$ and $\sigma_o$ have trained in their matches. The score related to units of type $u$ is given by $1 - \frac{|n_u - n'_u|}{\max(n_u, n'_u)}$. For example, if the number of Ranged units the strategy $p$ trained is 4 and if the number of Ranged units the strategy $\sigma_o$ trained is 10, then the score for Ranged units is $1 - 6/10 = 0.4$. The value returned is the average scores of all types of units and resources. The score is $1.0$ if both $p$ and $\sigma_o$ train the same number of units of each type and collect the same number of resources. We denote SA and UCT using this function as Sketch-SA(O) and Sketch-UCT(O). 

\subsection{Strategies to Clone}

We use weak and strong strategies $\sigma_o$ for generating the data sets $L$. 
$L$ is composed of state-action pairs from matches in which $\sigma_o$ plays the game with either itself or another strategy, which is specified below. For self-play matches, we include in $L$ only the state-action pairs of the winner.

\subsubsection{Can't Stop} We consider 3 data sets $L$, each generated with a different $\sigma_o$. The first $\sigma_o$ randomly chooses one of the available actions at each state of the game. 
The data set is composed of 3 self-play matches of this strategy, which wins approximately only 2.8\% of the matches it plays against $\sigma_{-i}$. We use a data set composed of 3 self-play matches of the GA strategy and a data set composed of 3 matches a human played with GA; the human player won all matches.

\subsubsection{MicroRTS} We also consider 3 data sets $L$ for MicroRTS. The first $L$ is composed of 2 matches (one in each starting location of the map) of Ranged Rush (RR), which is a simple programmatic strategy~\cite{stanescu2016evaluating}, against COAC. 
We also use a data set composed of 2 matches of A3N, a Monte Carlo tree search algorithm~\cite{moraes2018abstractions}, against COAC. A3N considers  low-level actions of units (e.g., move one square to the right) while planning their actions. We chose A3N because we reckon it would be hard for the synthesizer to clone its behavior as the strategies derived with A3N are unlikely to be in the space of strategies defined by the DSL (the DSL does not allow for a fine control of the units, as A3N does). Both RR and A3N are unable to win any matches against COAC, our $\sigma_{-i}$, in all maps evaluated. We also consider a data set composed of states from 2 self-play matches of COAC. 

\subsection{Empirical Results: Can't Stop}

\begin{figure}[t]
    \centering
    \includegraphics[width=.23\textwidth]{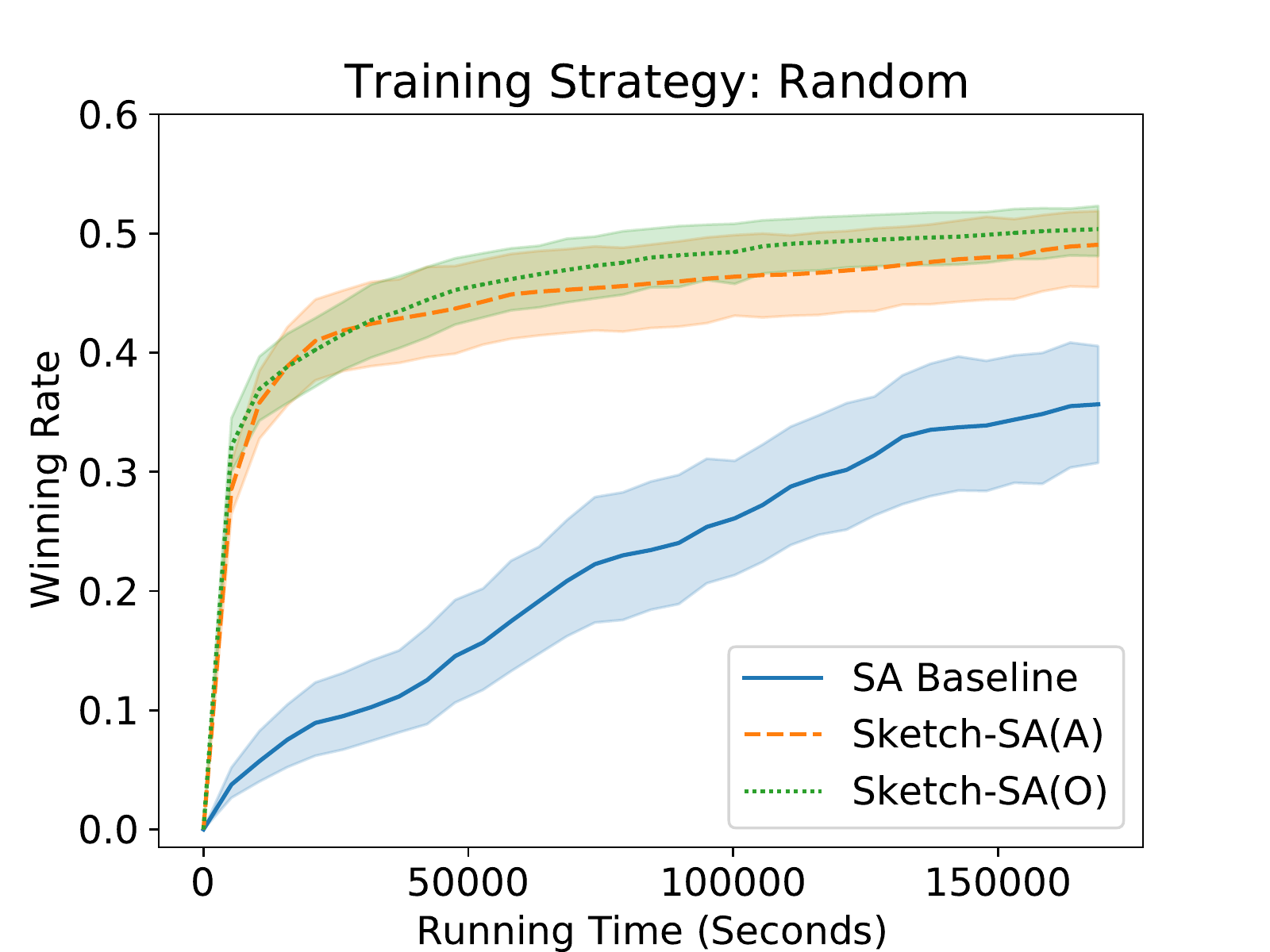}
    \includegraphics[width=.23\textwidth]{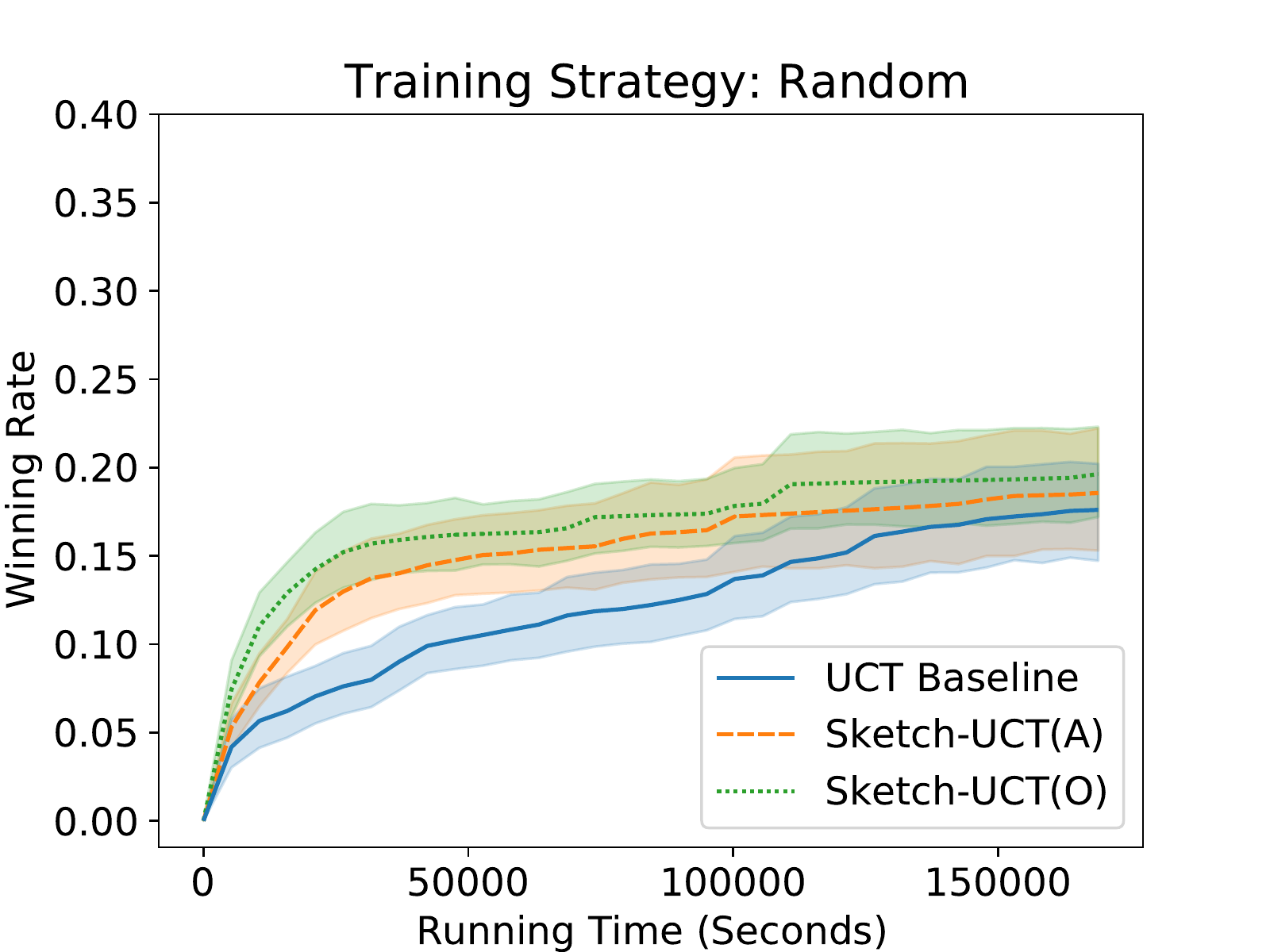}
    \includegraphics[width=.23\textwidth]{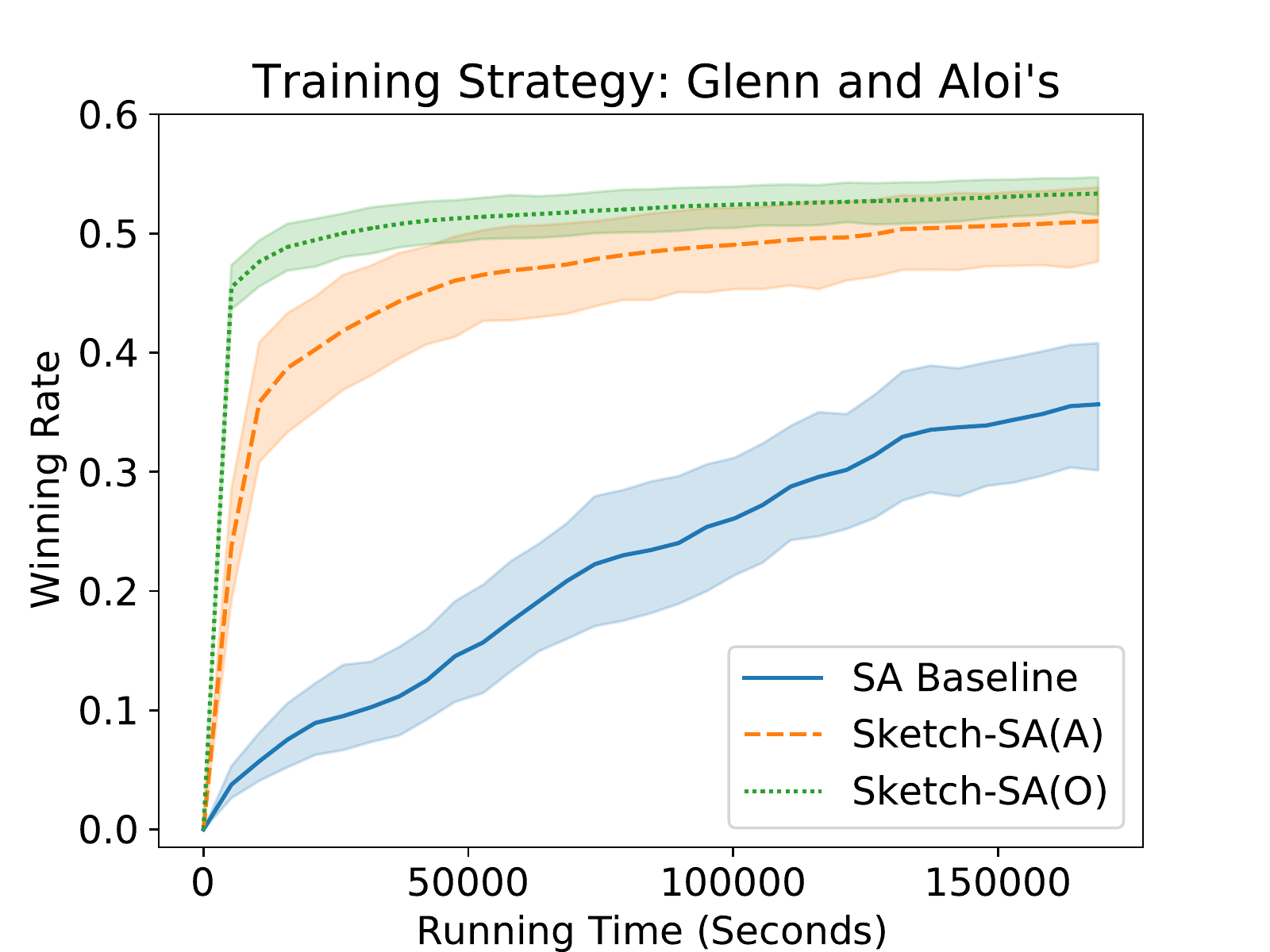}
    \includegraphics[width=.23\textwidth]{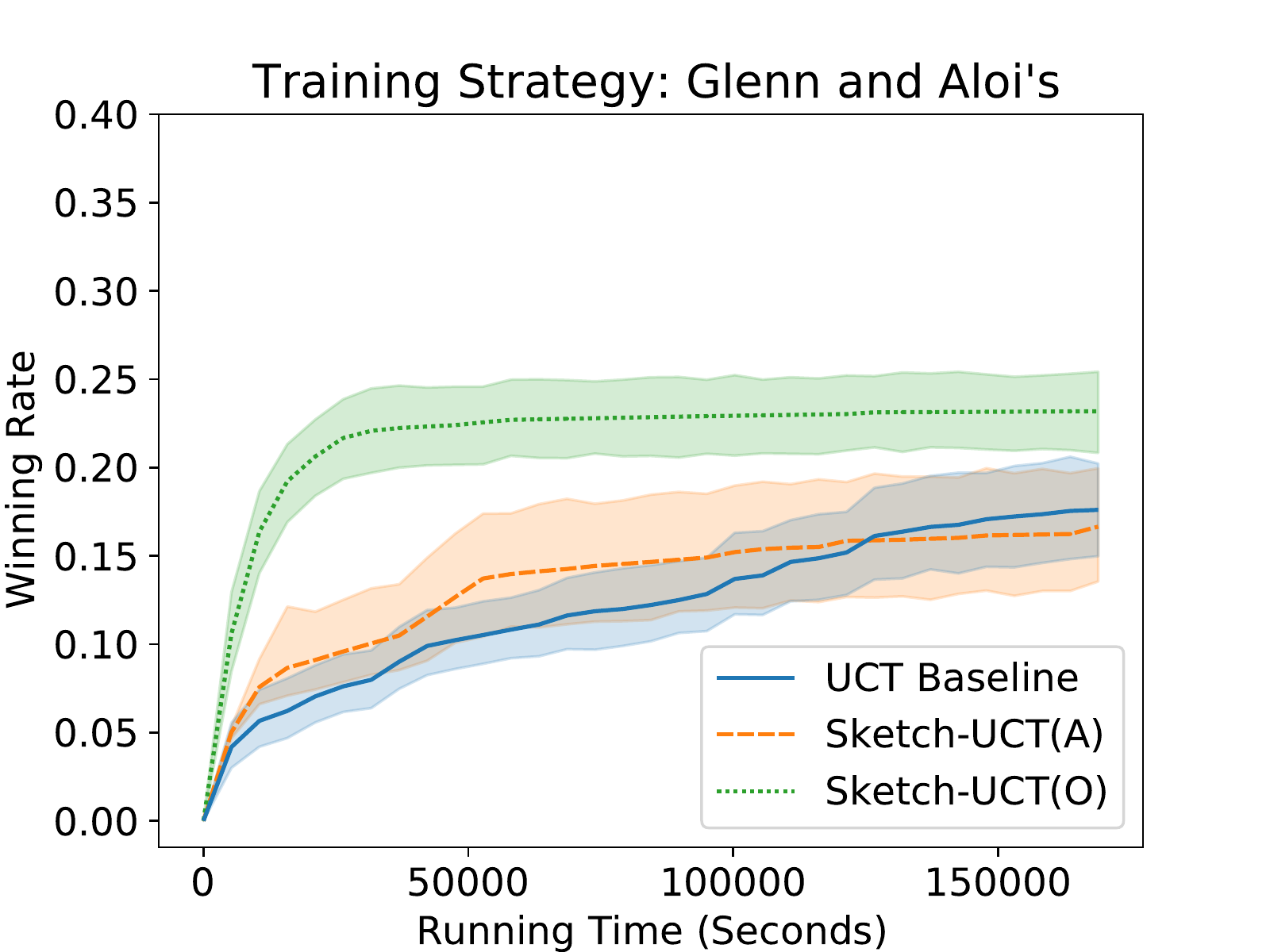}
    \includegraphics[width=.23\textwidth]{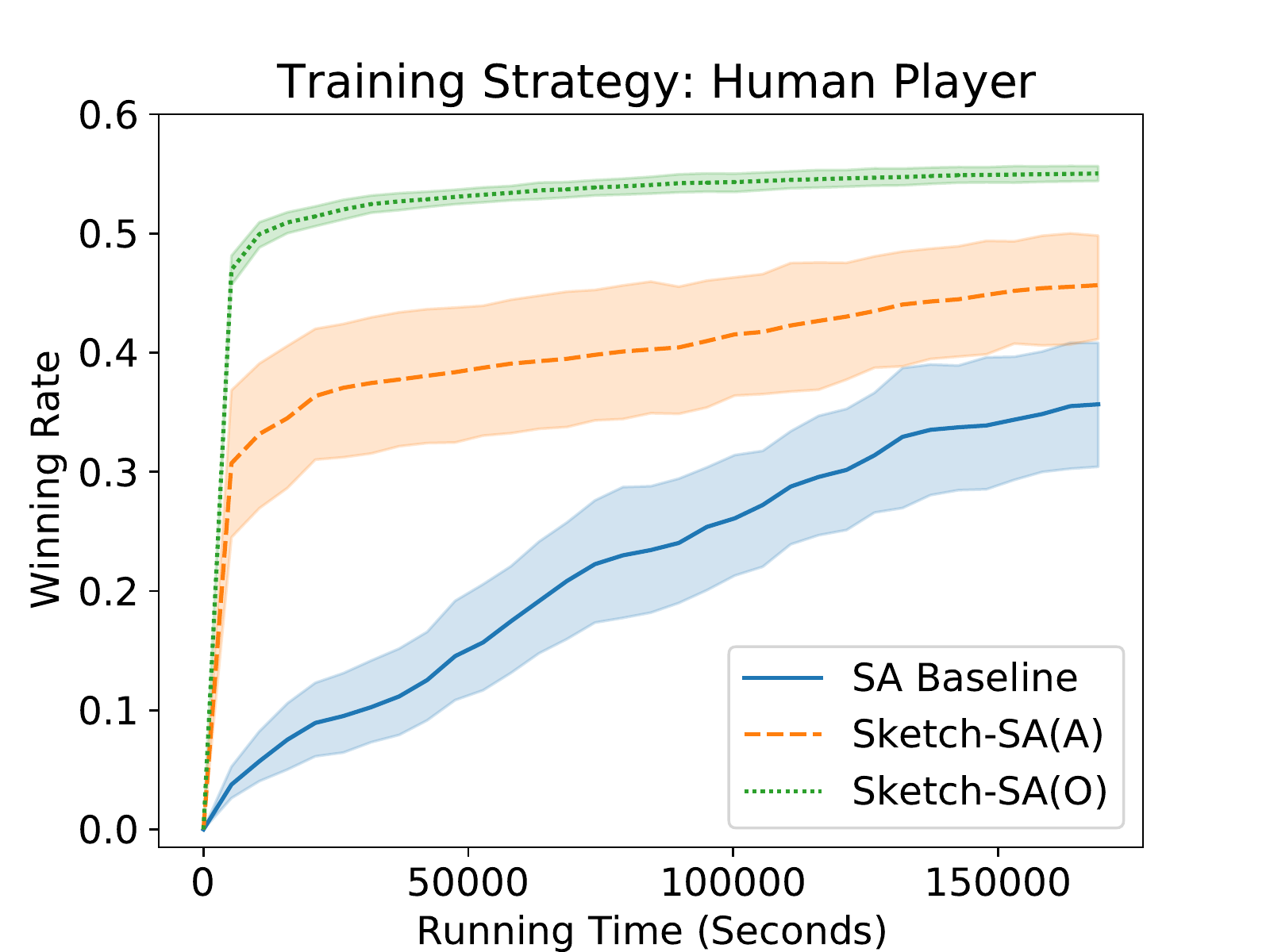}
    \includegraphics[width=.23\textwidth]{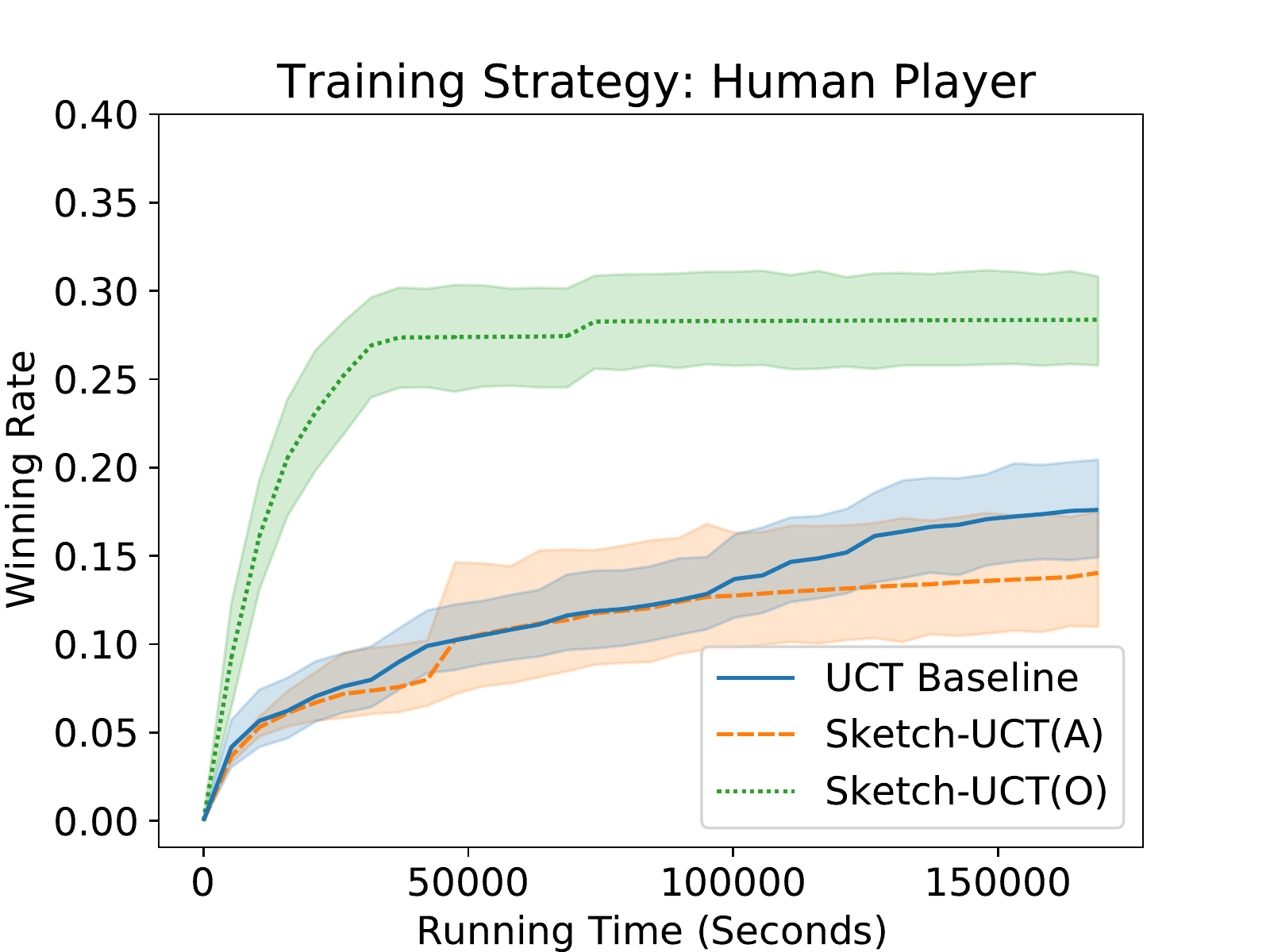}
    \caption{Winning rate of SA (left) and UCT (right) variants.}
    \label{fig:cantstop}
\end{figure}

\begin{figure*}[t]
    \centering
    \includegraphics[width=.24\textwidth]{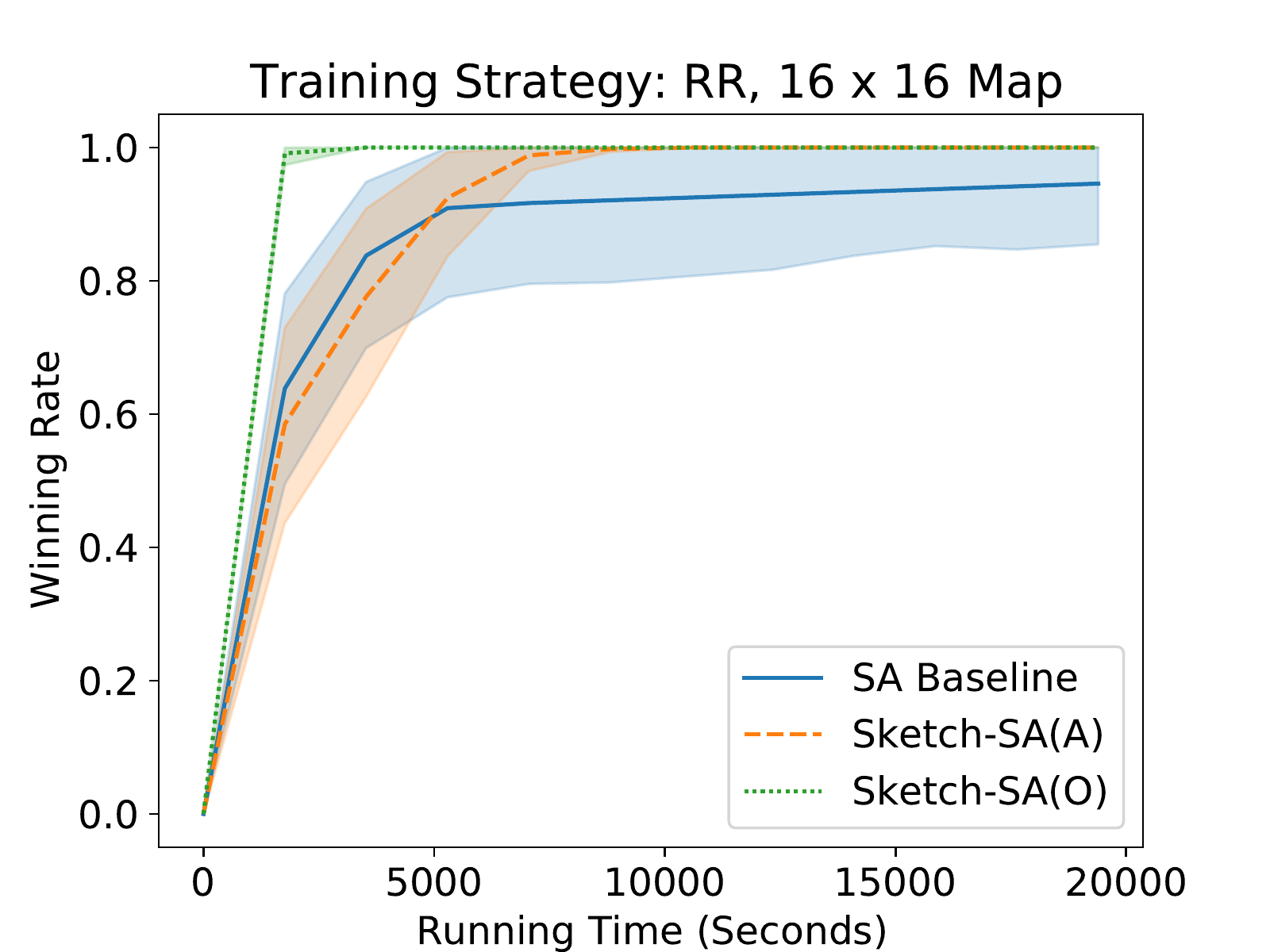}
    \includegraphics[width=.24\textwidth]{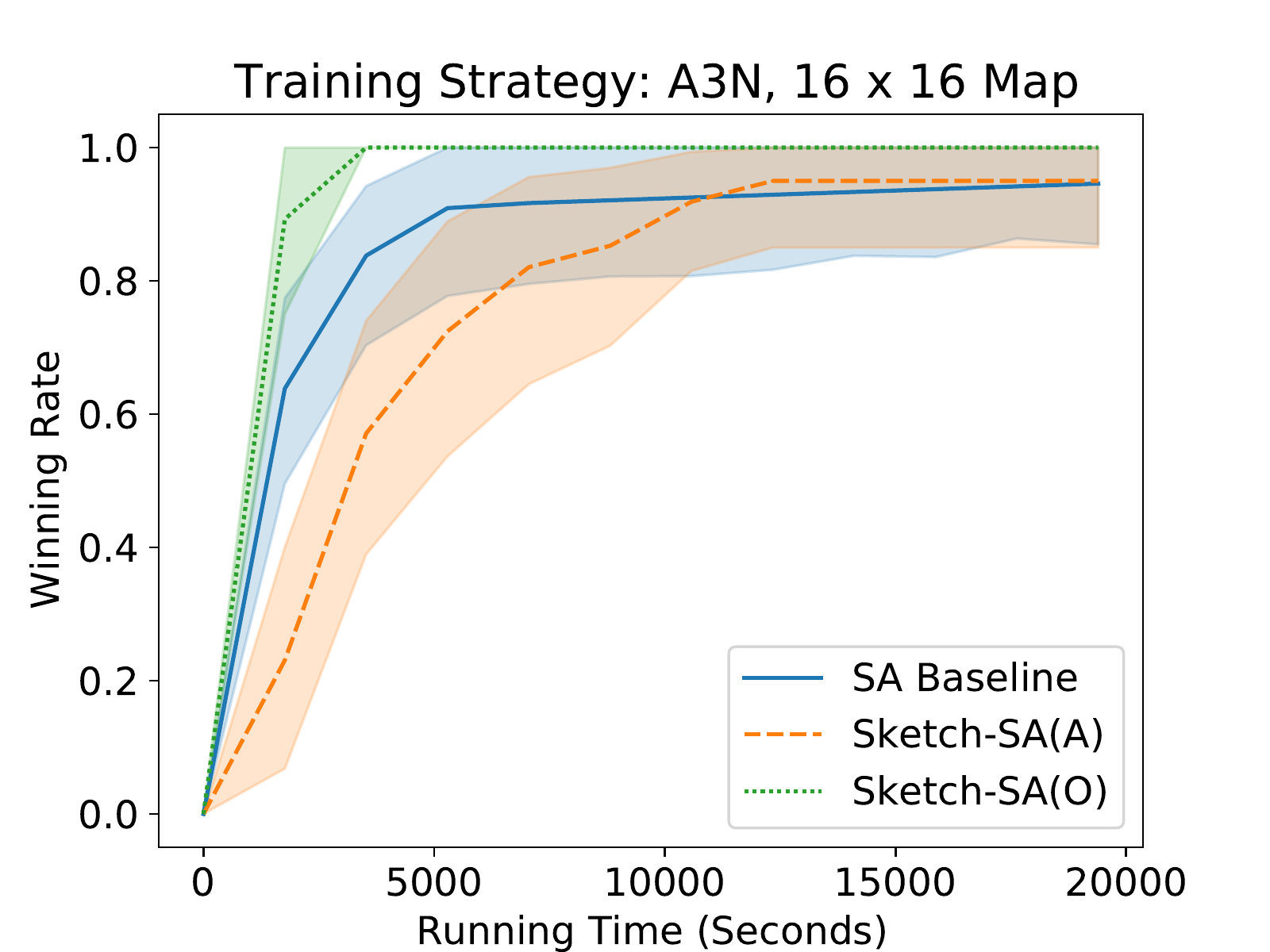}
    \includegraphics[width=.24\textwidth]{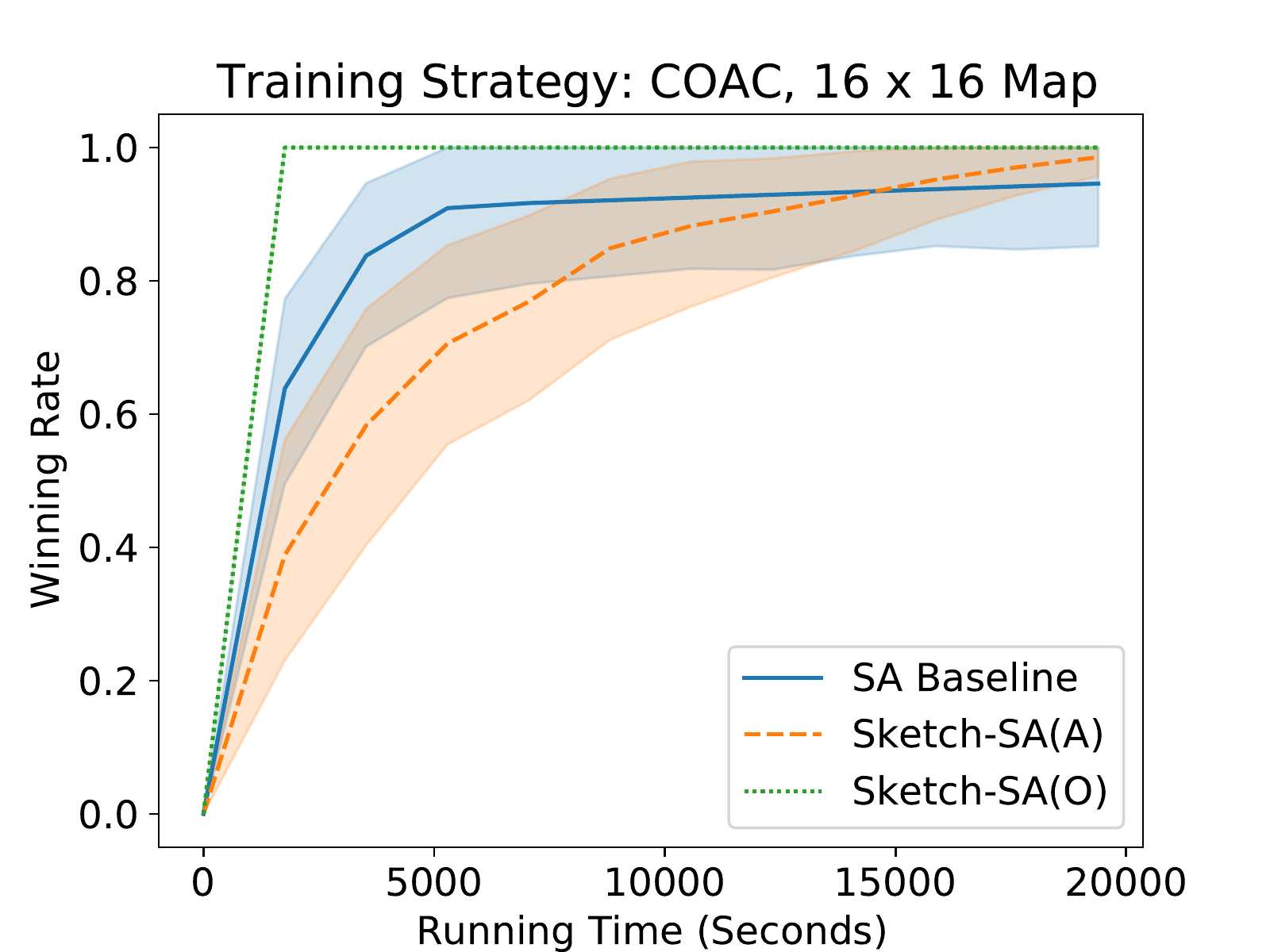}
    \includegraphics[width=.24\textwidth]{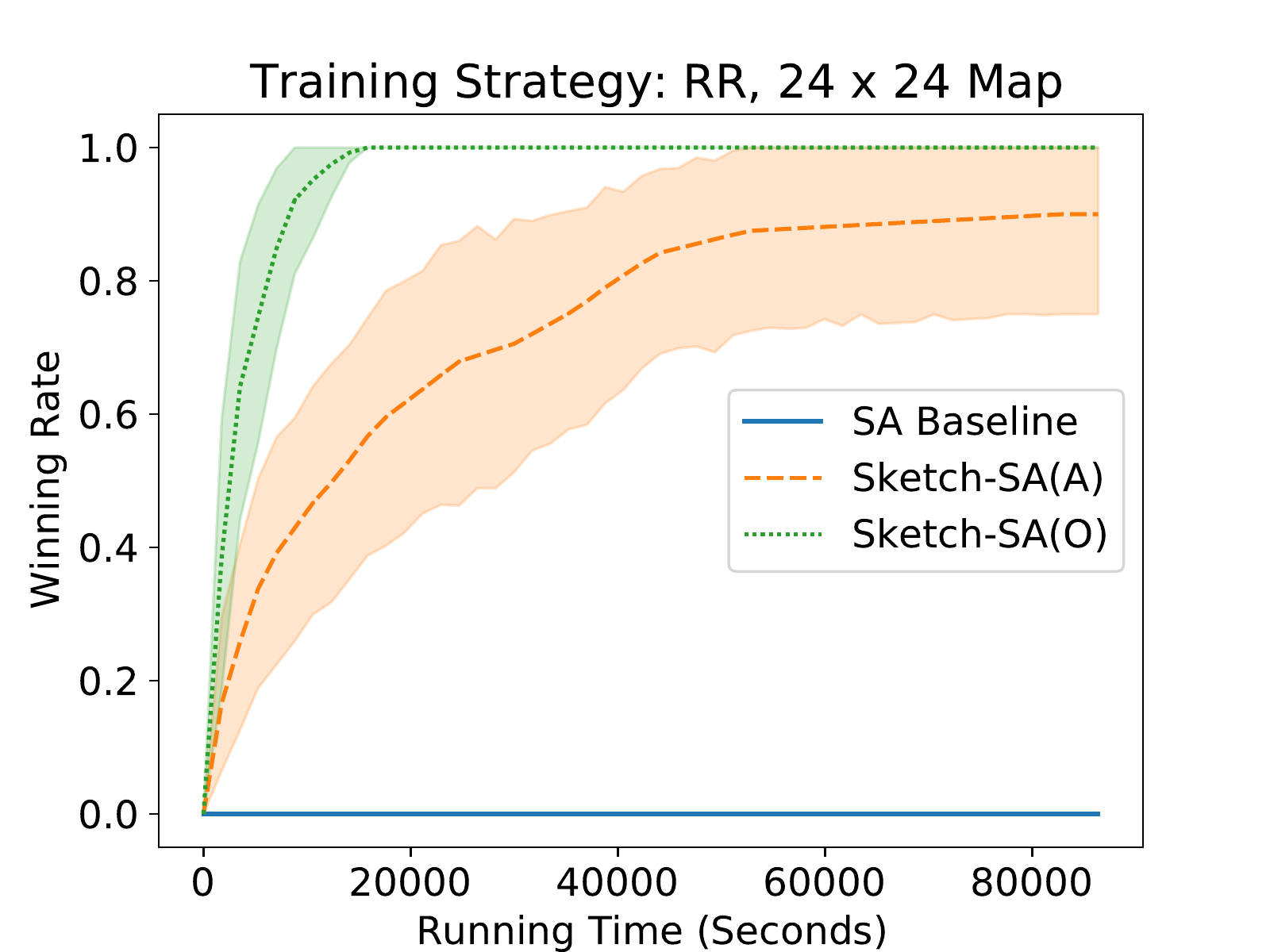}
    \includegraphics[width=.24\textwidth]{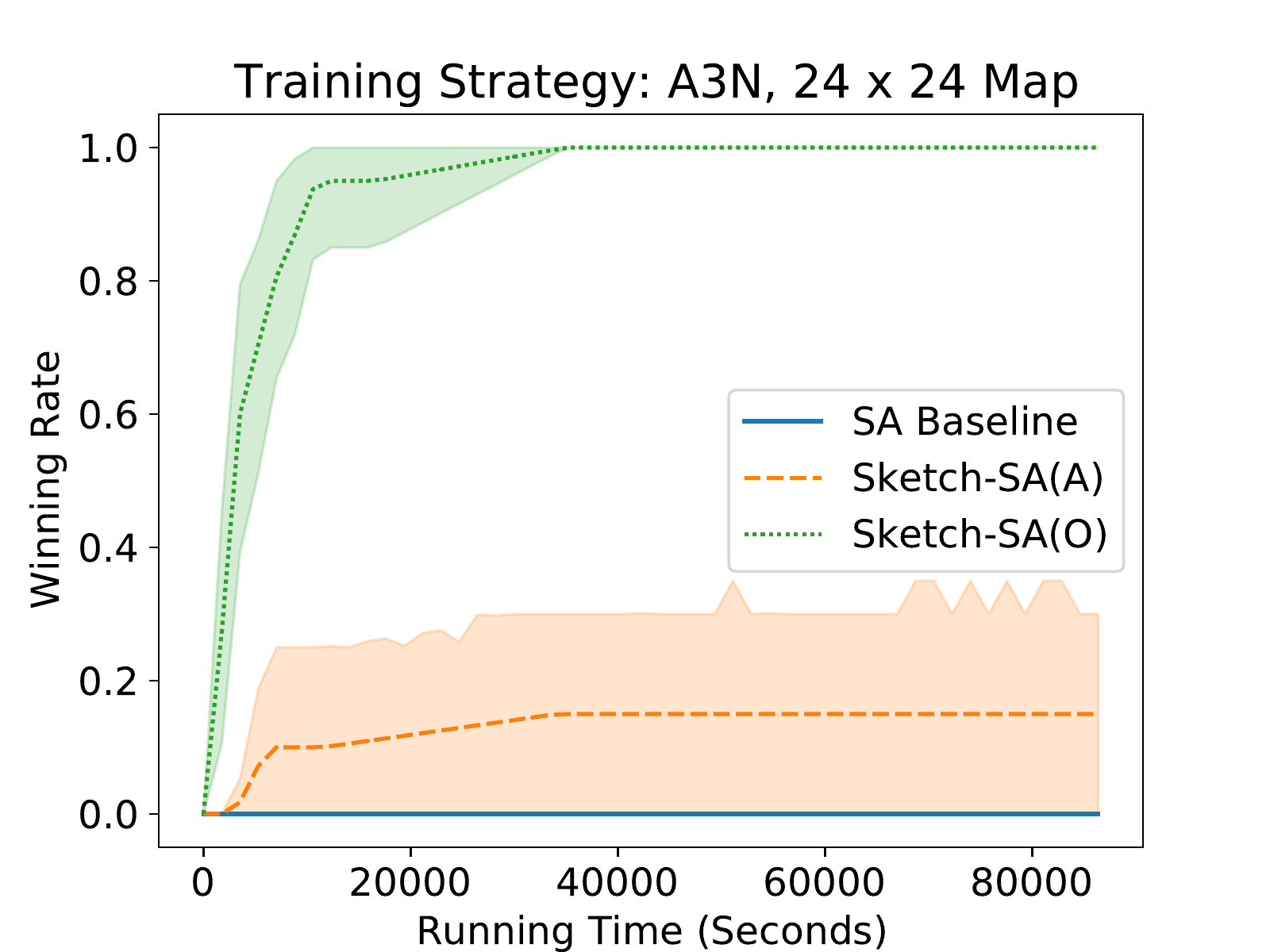}
    \includegraphics[width=.24\textwidth]{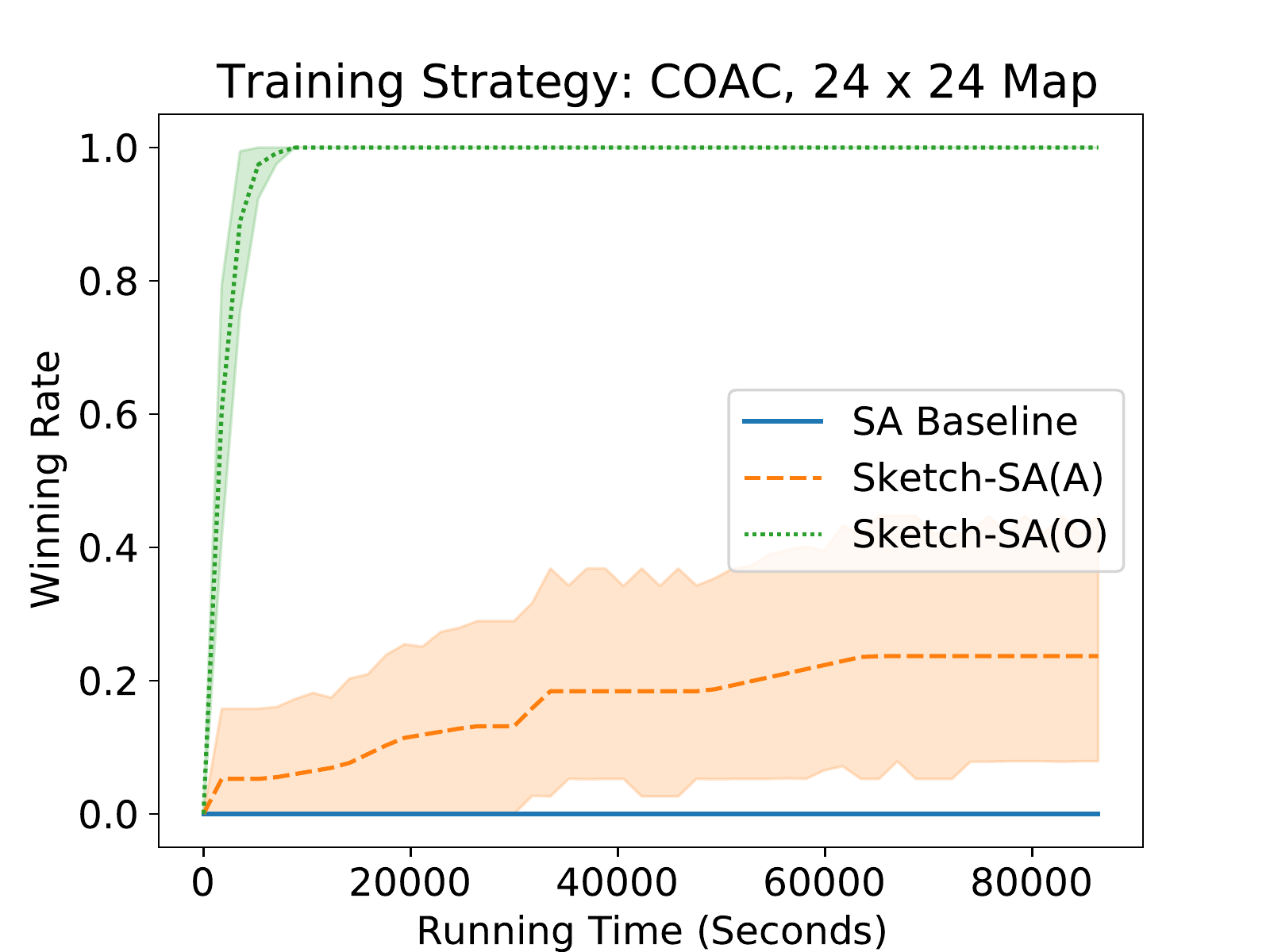}
    \includegraphics[width=.24\textwidth]{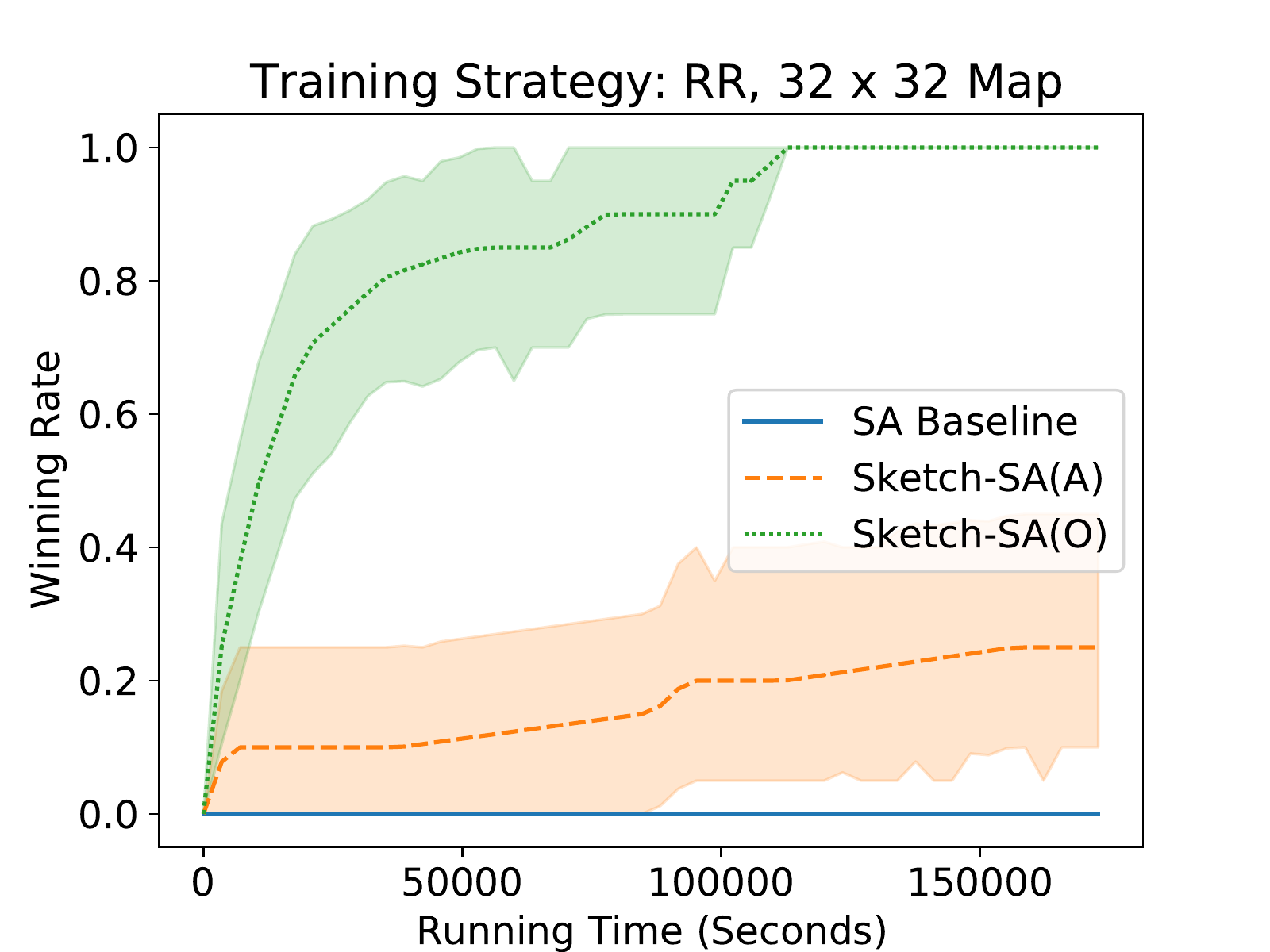}
    \includegraphics[width=.24\textwidth]{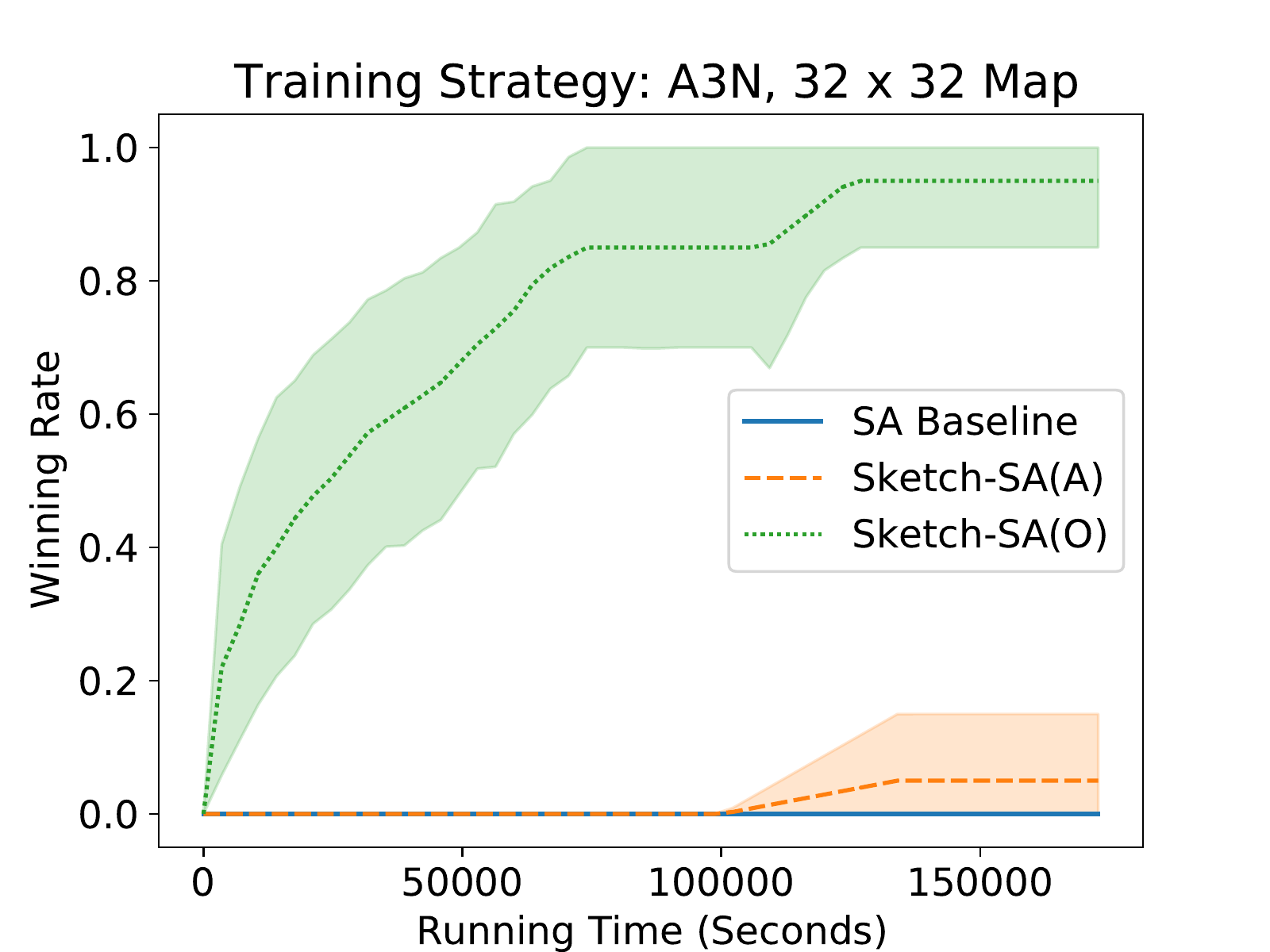}
    \includegraphics[width=.24\textwidth]{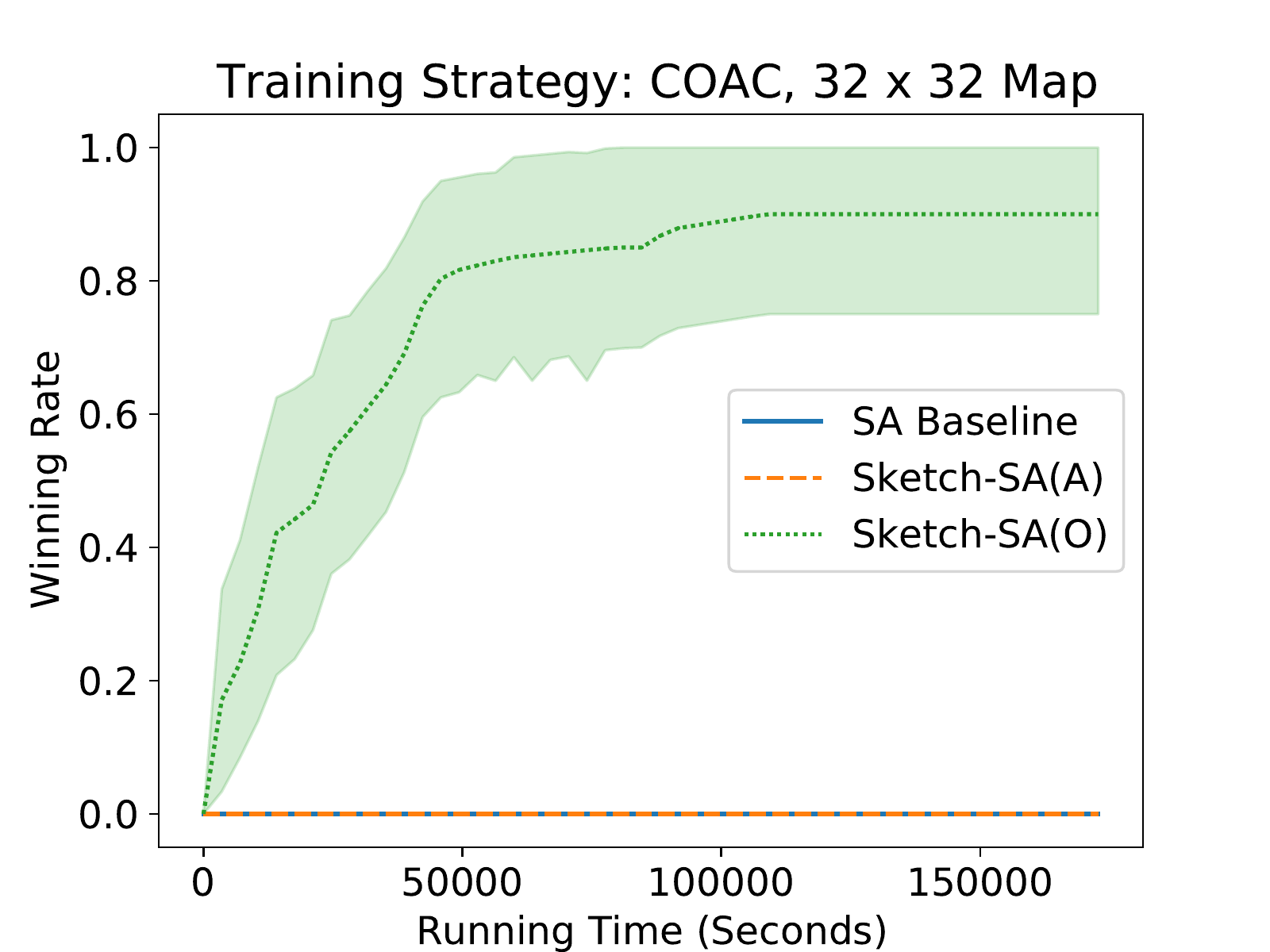}
    \includegraphics[width=.24\textwidth]{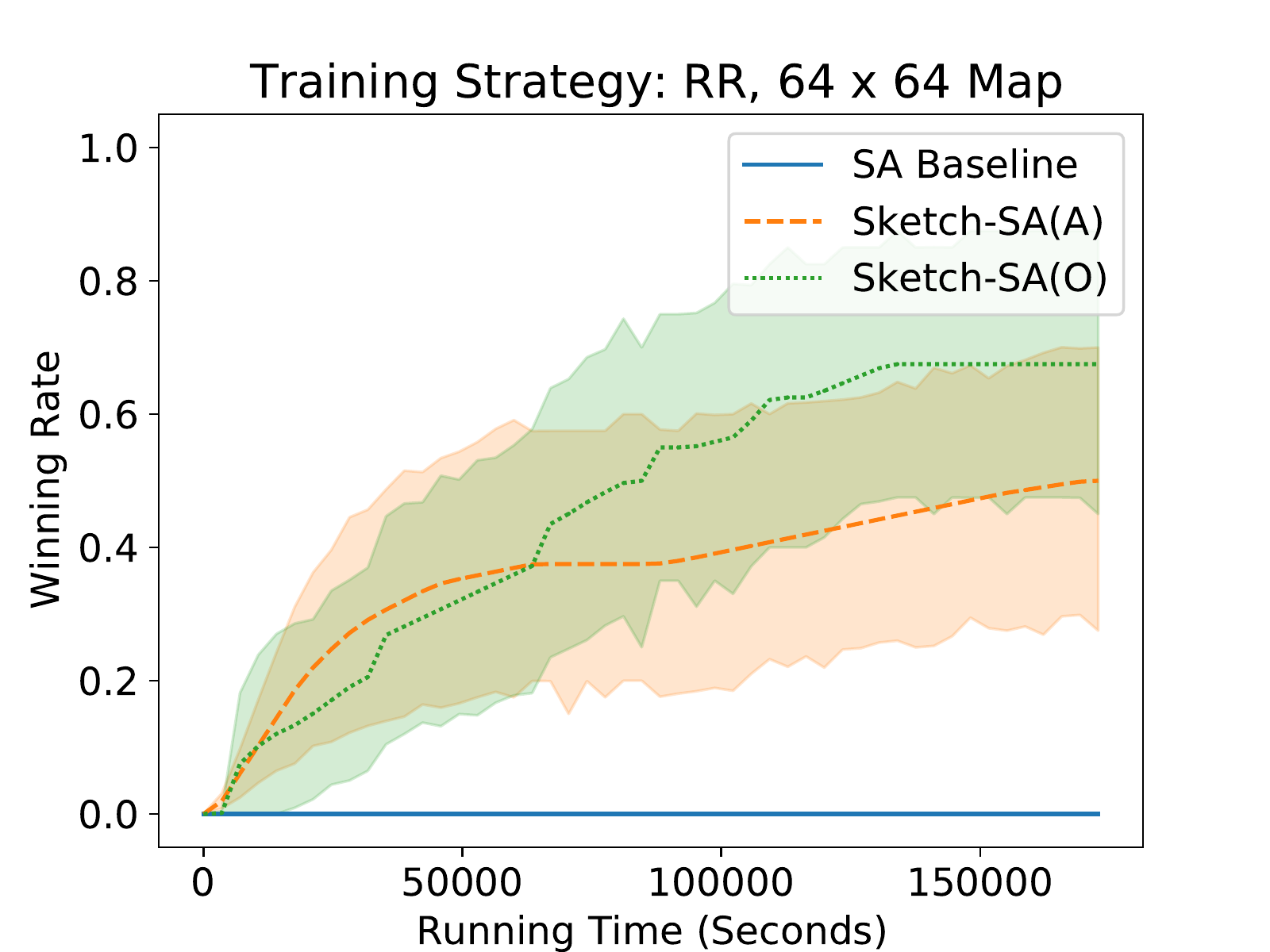}
    \includegraphics[width=.24\textwidth]{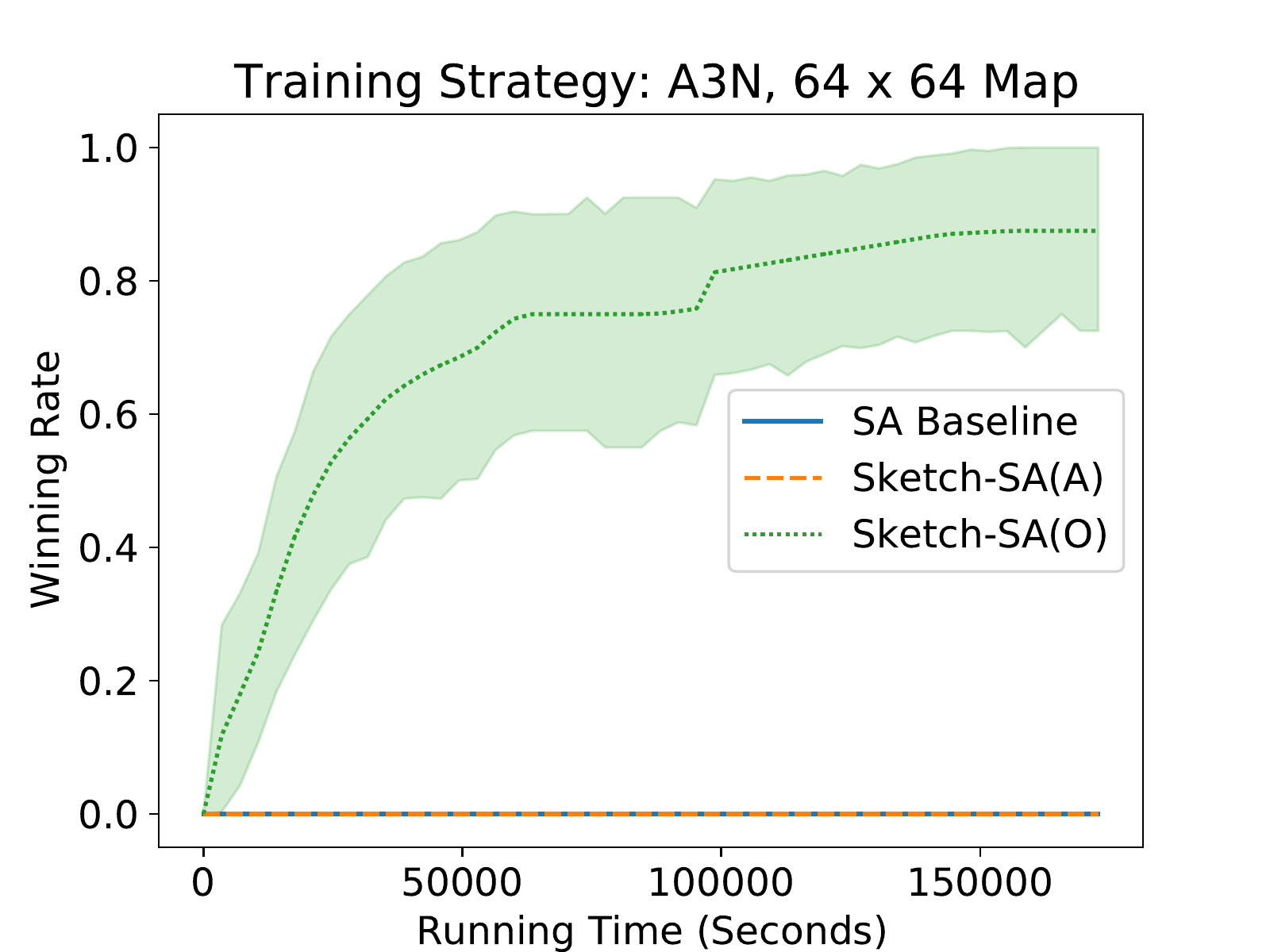}
    \includegraphics[width=.24\textwidth]{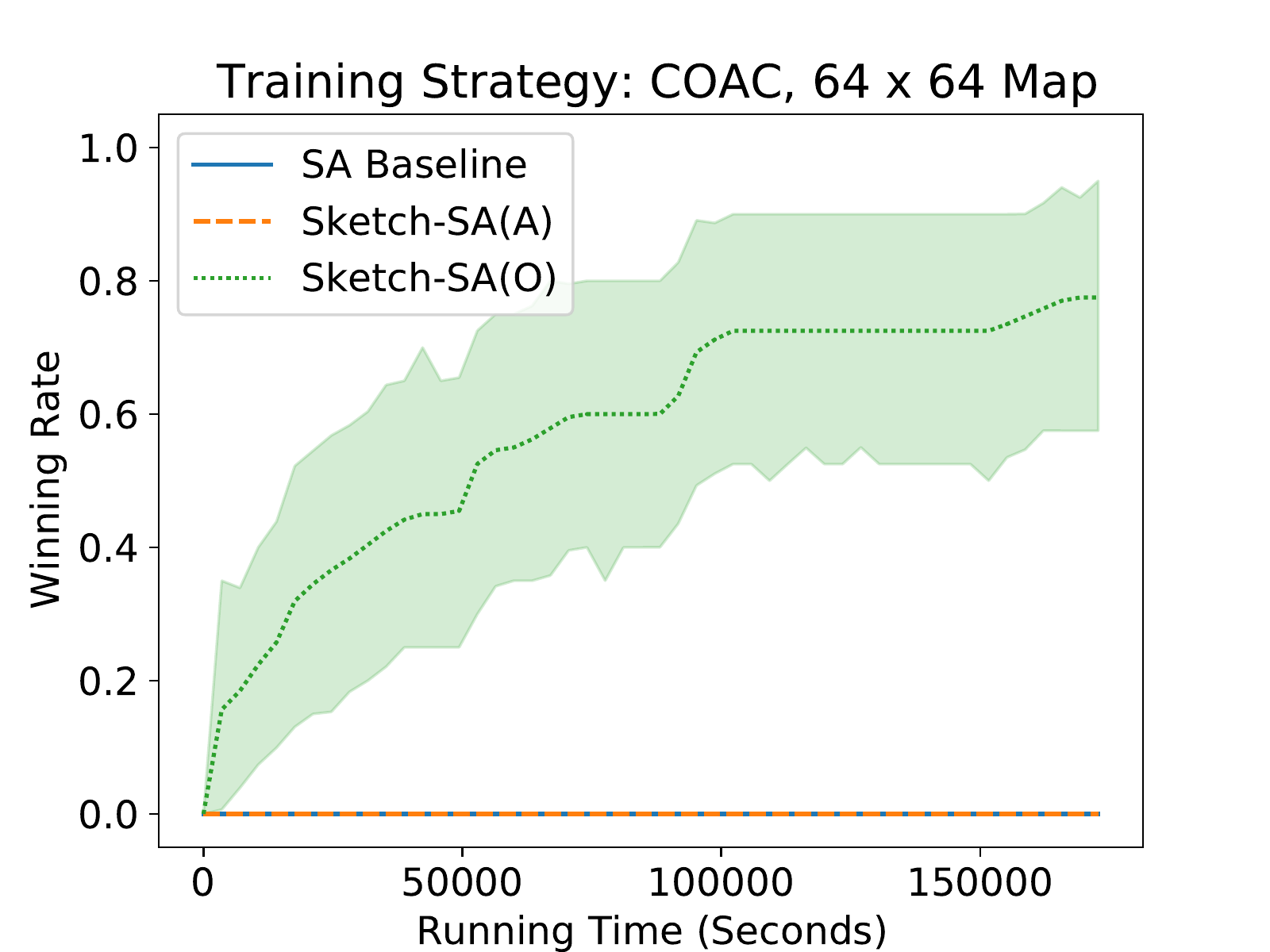}
    \caption{Winning rate of strategies SA variants synthesize; the maps increase in size from left to right and top to bottom.}
    \label{fig:microts}
\end{figure*}

Figure~\ref{fig:cantstop} shows the results on Can't Stop. Each plot shows the winning rate against $\sigma_{-i}$ (y-axis) of the best synthesized strategy over time (x-axis) of the methods: a search algorithm (either SA or UCT) without behavioral cloning for sketch learning (Baseline), a search algorithm that learns sketches: Sketch-SA(A), Sketch-UCT(A), Sketch-SA(O) or Sketch-UCT(O). In the plots we account for the time used in the sketch-searches. We ran 30 independent runs of each method and the lines represent the average results while the shaded areas the standard deviation of the runs. 

The learning-sketch methods are much faster than their baselines. In most of the cases, Sketch-SA and Sketch-UCT achieve winning rates that their baseline counterparts did not achieve within the time limit. The results also show that the SA methods perform better than their UCT counterparts. Only Sketch-SA synthesized strategies that defeat $\sigma_{-i}$ in more than 50\% of the matches. In particular, Sketch-SA(O) is the most effective method for approximating a best response for $\sigma_{-i}$. We conjecture SA synthesizes stronger strategies than UCT because it explores the space more quickly than UCT. The time complexity of UCT's selection step is quadratic on the program's length. This is because, in each selection step, the search traverses all production rules of the current incomplete program for each production rule applied to it. By contrast, SA can synthesize a large number of instructions with a single neighborhood operation. 

Although Sketch-SA(O) performs better by cloning the behavior of the human player, the method performs surprisingly well when it learns sketches by cloning the behavior of a random strategy. One of the key aspects for playing Can't Stop is to decide when to stop playing so that the neutral markers become permanent markers. The program must have a specific structure for computing the score that leads to a stop action, similar to \texttt{sum(map($\lambda.f$, neutrals))}. Here, \texttt{neutrals} is a list with the neutral markers and $f$ is a score function for individual markers. The \texttt{sum} and \texttt{map} operators return the sum of the scores of all markers. While the structure of this program is not trivial, the random strategy has a 50\% chance of choosing the stop action, and its effects are reflected on the states in $L$ (i.e., neutral markers become permanent markers). The synthesizers discover sketches like the program above because such programs place permanent markers on the board, as the random strategy does. The BR-search modifies the sketch to maximize the player's utility, but most of program's structure is maintained. 

The sketch-based methods perform worse with the action-based function. This is because the observation-based function captures the effects of even rare actions. A good player of Can't Stop chooses to continue playing in most states, but at crucial states they choose to stop. A player that never stops has a high action-based score because the stop action is rare. As a result, the sketch-based methods often fail to learn the program structure needed to correctly decide when to stop. 

\subsection{Empirical Results: MicroRTS}



Figure~\ref{fig:microts} shows the results of the SA variants on MicroRTS; we omit the UCT plots for space. The results of the UCT variants on MicroRTS are similar to those on Can't Stop. The UCT variants perform worse than their SA counterparts. In particular, no UCT method, including the baseline UCT, is able to synthesize a strategy that defeats $\sigma_{-i}$ on the larger 32$\times$32 and 64$\times$64 maps. Moreover, the approaches that learn sketches perform better than their counterparts on the smaller 16$\times$16 and 24$\times$24 maps. 
The plots for maps of size 16$\times$16 and 24$\times$24 in Figure~\ref{fig:microts} show a reduced running time (approximately 6 hours for the former and 24 hours for the latter) so we can better visualize the curves. Each line represents the average winning rate and the shaded areas the standard deviation of 10 independent runs of each method.

Like in Can't Stop, the sketch-based methods are superior to the baseline, with Sketch-SA(O) achieving winning rates near $1.0$ even when learning sketches from A3N, which is a strategy unable to defeat $\sigma_{-i}$. There is also a gap between the action and the observation-based functions and the gap seems to increase with the map size. Sketch-SA(A) did not synthesize strategies that defeated $\sigma_{-i}$ for $L$ generated with A3N and COAC on the 64$\times$64 map. The explanation for the poor performance of Sketch-SA(A) is similar to that on Can't Stop: some actions are rare but play an important role in the game (e.g., one can only train Ranged units after building a barracks, which might happen only once in a match). 

\subsection{Sample of Programmatic Strategy}

\begin{figure}[t!]
\begin{lstlisting}
def Sketch-SA-O-24x24(state s):
    for u in s:
        if not u.isWorker():
       	    u.moveToUnit(Ally, LessHealthy)
        u.train(Ranged)
        for u in s:
       	    u.attackIfInRange()
        u.build(Barracks)
    for u in s:
        u.harvest(4)
        u.attack(LessHealthy)
\end{lstlisting}
\caption{Sketch-SA(O)'s strategy for the 24$\times$24 map.}
\label{fig:sample_strategy}
\end{figure}

Figure~\ref{fig:sample_strategy} shows a strategy Sketch-SA(O) synthesized for the 24$\times$24 map. We lightly edited the strategy for readability. This strategy achieves the winning rate of $1.0$ against COAC. 
The strategy receives a state $s$ and assigns an action to each unit in $s$; if a unit is not assigned an action, then it does not perform an action in the next round of the game. Once the strategy assigns an action to a unit $u$, the action cannot be replaced by another action. For example, the strategy does not change the action assigned to units in line 4, even if we try to assign them a different action later in the program. This strategy trains Ranged units (line 5) once a barracks is built (line 8); a single barracks is built because all resources are spent training Ranged units once the barracks is available. The Ranged units cluster together (line 4) and attack enemy units within their range of attack (line 7). If there are no enemy units within their range, they attack the enemy's units that are close to being removed from the game (line 11). The strategy assigns 4 Workers to collect resources (line 10). This strategy 
is representative of the strategies our methods synthesize for both domains. 

\section{Conclusions}

In this paper we showed that behavioral cloning can be used to learn effective sketches for speeding up the synthesis of programmatic strategies. We presented Sketch-UCT and Sketch-SA, two synthesizers based on UCT and SA that learn a sketch for a program encoding an approximated best response to a target strategy by cloning the behavior of an existing strategy. The synthesizers use the sketch as a starting point in the search for an approximated best response. Experimental results on Can't Stop and MicroRTS showed that Sketch-SA can synthesize strategies able to defeat programmatic strategies written by human programmers in all settings tested, even when learning sketches from weak strategies. In particular, Sketch-SA synthesized strategies that defeated the winner of the latest MicroRTS competition on all maps used in our experiments, while baseline synthesizers failed to generate good strategies in these settings.  

\section{Acknowledgements} This research was partially supported by FAPEMIG, CAPES, and Canada's CIFAR AI Chairs program. The research was carried out using computational resources from Compute Canada. We thank the anonymous reviewers and Vadim Bulitko for their feedback. 

\bibliography{references}

\appendix

\section{BR-search UCT Tree Initialization}
 
\begin{figure}[t]
  \begin{center}
\includegraphics[width=230px]{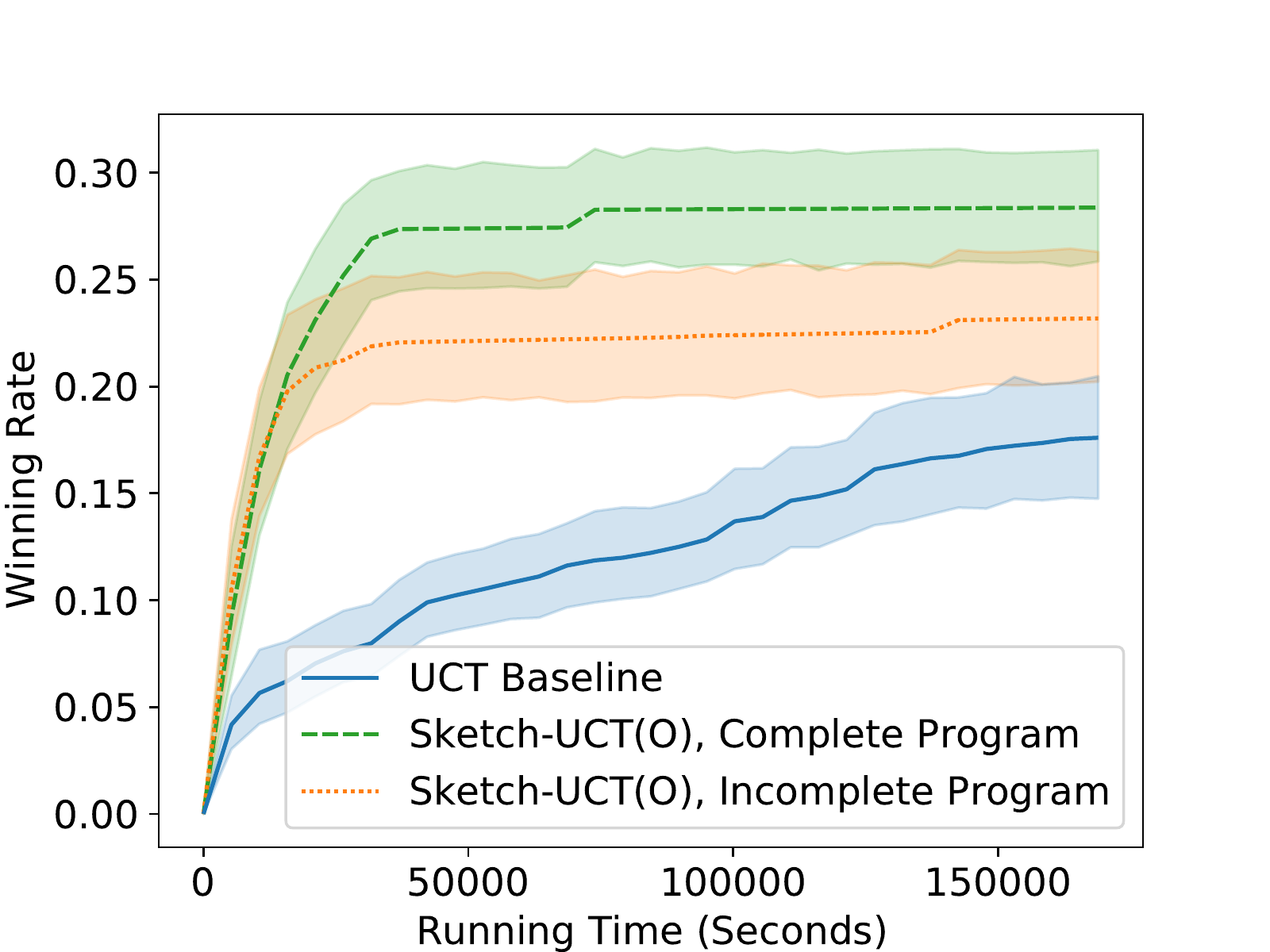}
\end{center}
\caption{Methods of initialization of the BR-search tree}
\label{fig:incomplete_vs_complete_uct}
\end{figure}

 
 We considered two approaches to initialize the UCT tree of the BR-search: we can use the complete program $p$ or the incomplete program given by the production rules represented by nodes in the UCT tree of the sketch-search. We empirically observed that initializing the program with the complete program sped up the synthesis step during the BR-search, as one can observe in Figure \ref{fig:incomplete_vs_complete_uct}.
 
 
\section{Domain-Specific Language for Can't Stop}

The DSL we developed for Can't Stop is described by the following context-free grammar. 

\begin{align*}
S &\to AA \\
A &\to \text{if}(B < B) \text{ then } A \text{ else } A\, \GOR \text{argmax}(L) \GOR \text{sum}(L) \\ 
&\to E \otimes E \\
B &\to N \GOR E \otimes E \\ 
E &\to E \otimes E \GOR N \GOR  \text{sum}(L) \GOR L_2 \GOR F_2\\ 
L &\to \text{map}(\lambda F_1,L) \GOR L_1 \GOR l_1 \\
\lambda F_1 &\to \text{sum}(L) \GOR \text{map}(\lambda F_1,L)  \GOR E \otimes E \\ 
F_2 &\to f_1 \GOR f_2  \GOR f_3 \GOR f_4 \GOR f_5 \GOR f_6\\ 
L_1 &\to l_2 \GOR l_3 \\
L_2 &\to l_4 \GOR l_5 \\
N &\to 0 \GOR 1 \\ 
\otimes &\to + \GOR - \GOR * \\ 
\end{align*}

This DSL allows our system to synthesize actions for both decisions in Can't Stop: yes-no decision and column decision. 
$\lambda F_1$ represents lambda functions used as parameters for the \texttt{map} operator and $F_2$ is the set of domain-specific functions that provide information about a state of the game. Next, we describe these functions; in parenthesis we show the name of the function used in our codebase. 
\begin{itemize}
\item\emph{$f_1$}: Calculates the number of cells the player has advanced in the current round (NumberAdvancedThisRound).
\item\emph{$f_2$}: Calculates the number of cells the player will advance if they take the action passed as a parameter (NumberAdvancedByAction).
\item\emph{$f_3$}:  Calculates how many cells the player has secured in the column passed as parameter (PositionsPlayerHasSecuredInColumn).
\item\emph{$f_4$}: Calculates how many cells the opponent has secured in the column passed as parameter (PositionsOpponentHasSecuredInColumn).
\item\emph{$f_5$}: Calculates a difficulty score \cite{Glenn2009AGH} of a state, where the value returned depends on the position of the neutral tokens; a state is considered more difficult if the neutral tokens are on odd columns or if all tokens are either on columns with numbers smaller than 7 or greater than 7
(DifficultyScore).
\item\emph{$f_6$}: Returns 1 if the action uses a neutral token, returns 0 otherwise (IsNewNeutral).

\end{itemize}


\begin{itemize}

\item\emph{$l_1$}: The set of actions is represented as a list of lists. When used in a \texttt{map} operator, we call the lists inside an action list a ``locallist,'' which is represented in the DSL as $l_1$.
\item\emph{$l_2$}: List that represents which columns the neutral tokens are located given the state passed as parameter (neutrals). 
\item\emph{$l_3$}:  List that represents the actions available for the state passed as parameter (actions).
\item\emph{$l_4$}: List that represents weights of the columns used in Glenn and Aloi's strategy. In their strategy this set of weights is used in the yes-no decision. In our DSL, this list can be used by both actions. $l_4 = [7, 7, 3, 2, 2, 1, 2, 2, 3, 7, 7]$ (progress\_value).
\item\emph{$l_5$}: List that represents weights of the columns used in Glenn and Aloi's strategy in the column decision. In our DSL, this list can be used in both decisions. $l_5 = [7, 0, 2, 0, 4, 3, 4, 0, 2, 0, 7]$ (move\_value).

\end{itemize}

\subsection{Domain-Specific Language for MicroRTS}

Our DSL for MicroRTS is inspired on that of \citet{Marino2021}. The DSL is described by the following context-free grammar.

\begin{align*}
S &\to SS \GOR \text{for S} \GOR \text{if}(B) \text{ then } S \GOR \text{if}(B) \text{ then } S \text{ else } S\\
&\to C \GOR \lambda \\
B &\to f_1(T, N) \GOR f_2(T, N)  \GOR f_3(T, N) \GOR f_4(N) \GOR f_5(N)  \\
&\to \GOR f_6(N) \GOR f_7(T) \GOR f_8 \GOR f_9 \GOR f_{10} \GOR f_{11} \GOR f_{12} \GOR f_{13} \GOR f_{14} \\
C &\to c_1(T, D, N) \GOR c_2(T, D, N)  \GOR c_3(T_p, O_p) \GOR c_4(O_p)  \\
&\to c_5(N) \GOR c_6 \GOR c_7 \\
T &\to \text{Base} \GOR \text{Barracks} \GOR \text{Ranged} \GOR \text{Heavy} \GOR \text{Light} \\
&\to \text{Worker} \\
N &\to 0 \GOR 1 \GOR 2 \GOR 3 \GOR 4 \GOR 5 \GOR 6 \GOR 7 \GOR 8 \GOR 9 \GOR 10 \GOR 15 \GOR 20 \GOR 25 \\
&\to 50 \GOR 100 \\
D &\to \text{EnemyDir} \GOR \text{Up} \GOR \text{Down} \GOR \text{Right} \GOR \text{Left} \\ 
O_p &\to \text{Strongest} \GOR \text{Weakest} \GOR \text{Closest} \GOR \text{Farthest}  \\
&\to \text{LessHealthy} \GOR \text{MostHealthy} \GOR \text{Random} \\
T_p &\to \text{Ally} \GOR \text{Enemy} \\ 
\end{align*}

This DSL allows nested loops and conditionals. It contains several Boolean functions (B) and command-oriented functions (C) that provide either information about the current state of the game or commands for the ally units.

Next, we describe the Boolean functions used in our DSL.

\begin{itemize}
\item\emph{$f_1(T,N)$}: Checks if the ally player has $N$ units of type $T$ (HasNumberOfUnits).
\item\emph{$f_2(T,N)$}: Checks if the opponent player has $N$ units of type $T$ (OpponentHasNumberOfUnits).
\item\emph{$f_3(T,N)$}:  Checks if the ally player has less than $N$ units of type $T$ (HasLessNumberOfUnits).
\item\emph{$f_4(N)$}: Checks if the ally player has $N$ units attacking the opponent (HaveQtdUnitsAttacking).
\item\emph{$f_5(N)$}: Checks if the ally player has a unit within a distance $N$ from a opponent's unit (HasUnitWithinDistanceFromOpponent).
\item\emph{$f_6(N)$}: Checks if the ally player has $N$ units of type Worker harvesting resources (HasNumberOfWorkersHarvesting).
\item\emph{$f_7(T)$}: Checks if a unit is an instance of  Type $T$ (is\_Type). 
\item\emph{$f_8$}: Check if a unit is of type Worker (IsBuilder).
\item\emph{$f_9$}: Checks if a unit can attack (CanAttack).
\item\emph{$f_{10}$}: Checks if the ally player has a unit that kills an opponent's unit with one attack action (HasUnitThatKillsInOneAttack).
\item\emph{$f_{11}$}: Checks if the opponent player has a unit that kills an ally's unit with one attack action (OpponentHasUnitThatKillsUnitInOneAttack).
\item\emph{$f_{12}$}: Checks if an unit of the ally player is within attack range of an opponent's unit (HasUnitInOpponentRange).
\item\emph{$f_{13}$}: Checks if an unit of the opponent player is within attack range of an ally's unit (OpponentHasUnitInPlayerRange).
\item\emph{$f_{14}$}: Checks if a unit can harvest resources (CanHarvest).
\end{itemize}

Next, we describe the command-oriented functions used in our DSL:

\begin{itemize}
\item\emph{$c_1(T,D,N)$}: Trains $N$ units of type $T$ on a cell located on the $D$ direction of the unit (Build).
\item\emph{$c_2(T,D,N)$}: Trains $N$ units of type $T$ on a cell located on the $D$ direction of the structure responsible for training them (Train).
\item\emph{$c_3(T_p,O_p)$}:  Commands a unit to move towards the player $T_p$ following a criterion $O_p$ (moveToUnit).
\item\emph{$c_4(O_p)$}: Commands a unit to attack units of the opponent player following a criterion $O_p$ (Attack).
\item\emph{$c_5(N)$}: Sends $N$ Worker units to harvest resources (Harvest).
\item\emph{$c_6$}: Commands a unit to stay idle and attack if an opponent units comes within its attack range (Idle).
\item\emph{$c_7$}: Commands a unit to move in the opposite direction of the player's base (MoveAway).
\end{itemize}

$T$ represents the types a unit can assume. $N$ is a set of integers. $D$ represents the directions available used in action functions. $O_p$ is a set of criteria to select an opponent unit based on their current state. $T_p$ represents the set of players. See the different types, integers, directions, criteria, and players we used in the context-free grammar above. 

\section{UCT Results for MicroRTS}

\begin{figure*}[t]
    \centering
    \includegraphics[width=.24\textwidth]{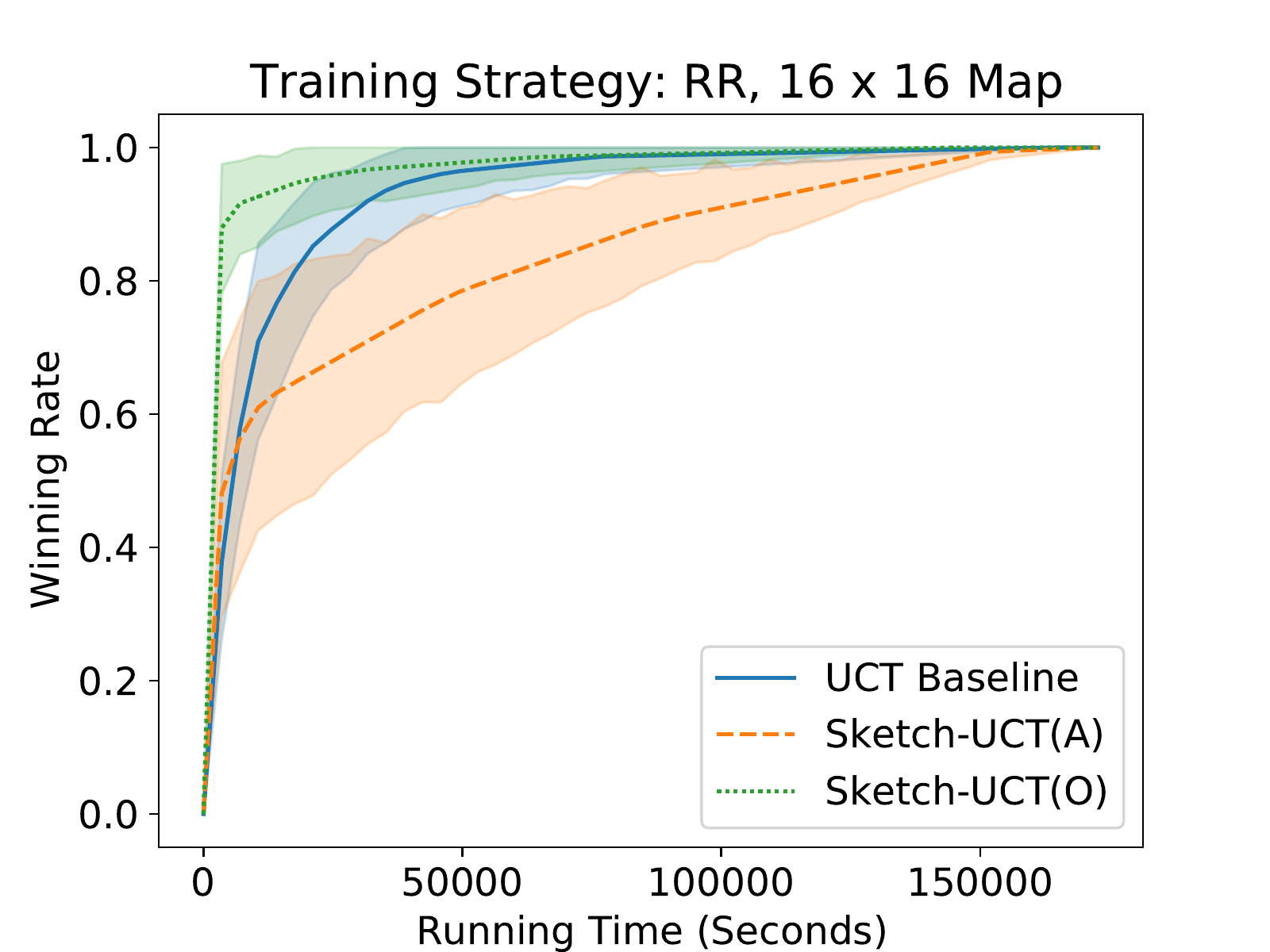}
    \includegraphics[width=.24\textwidth]{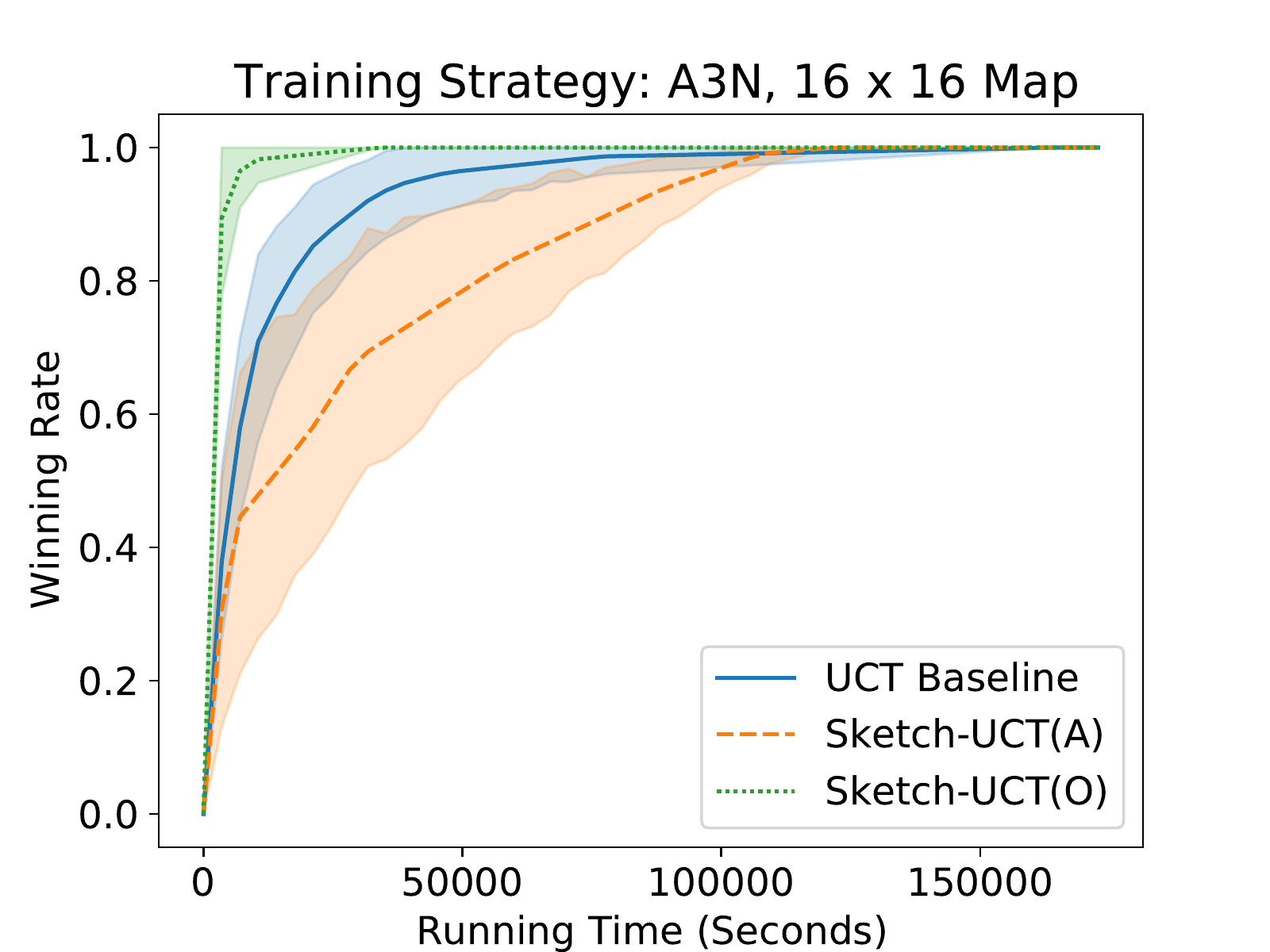}
    \includegraphics[width=.24\textwidth]{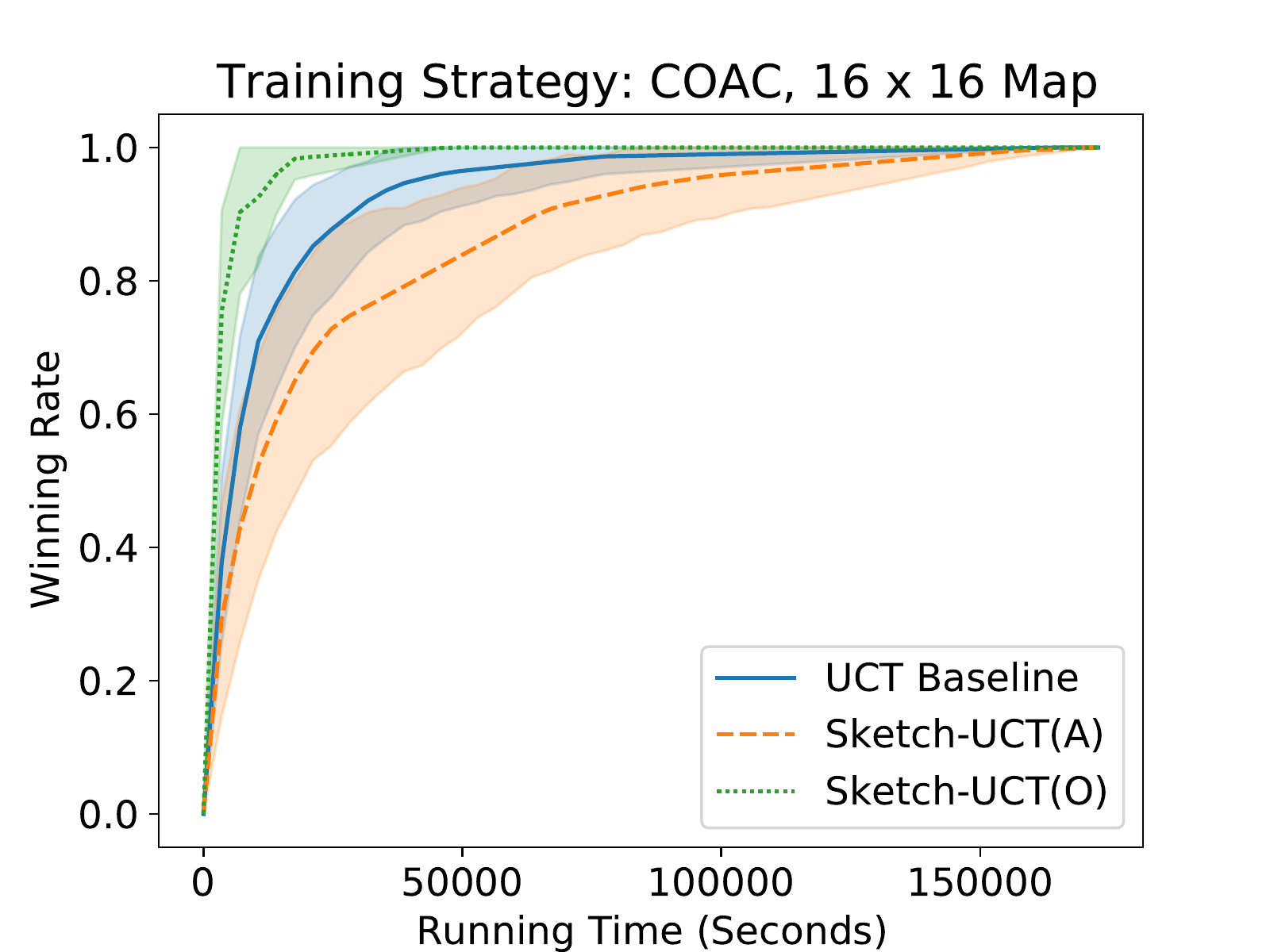}
    \includegraphics[width=.24\textwidth]{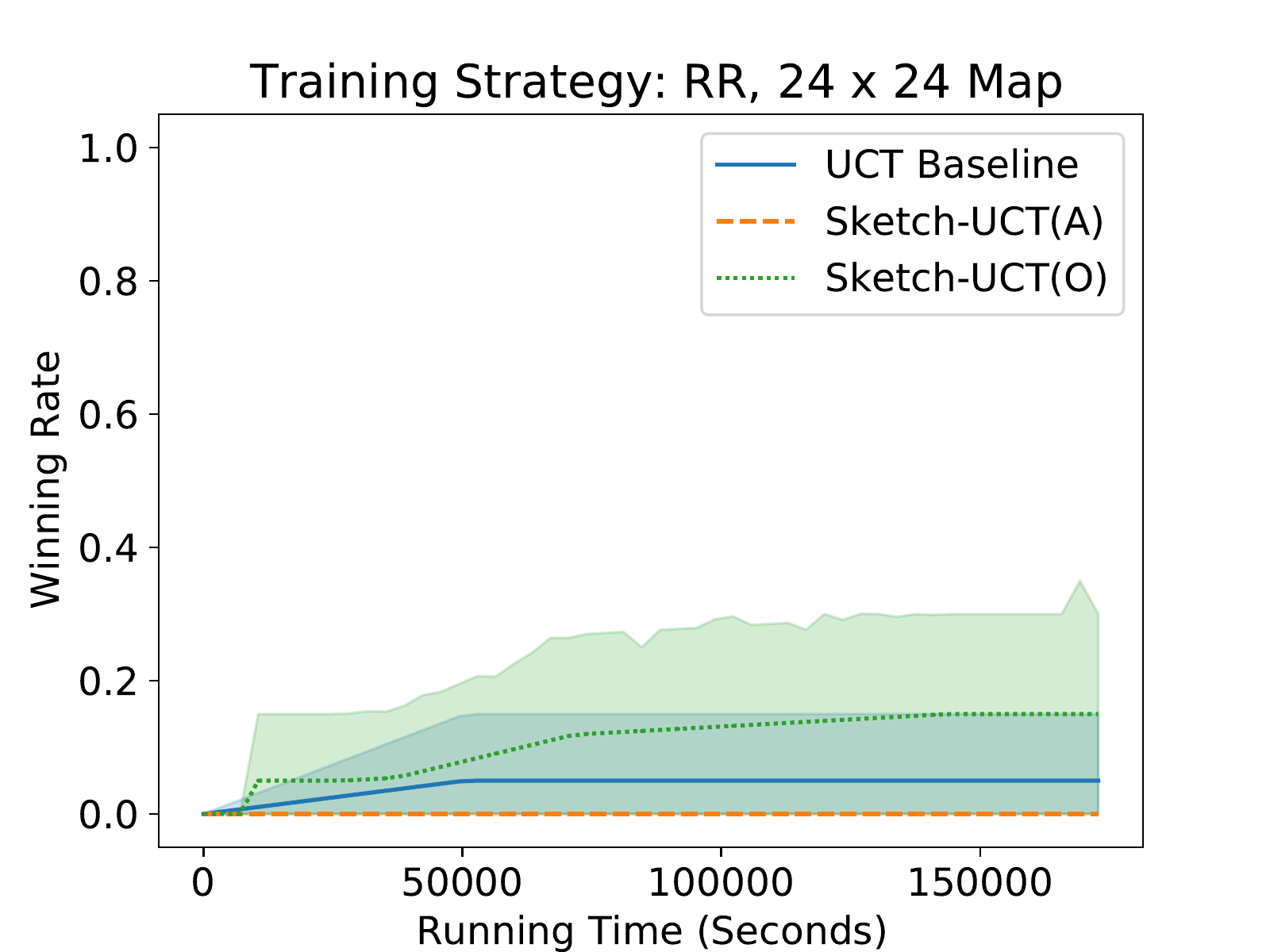}
    \includegraphics[width=.24\textwidth]{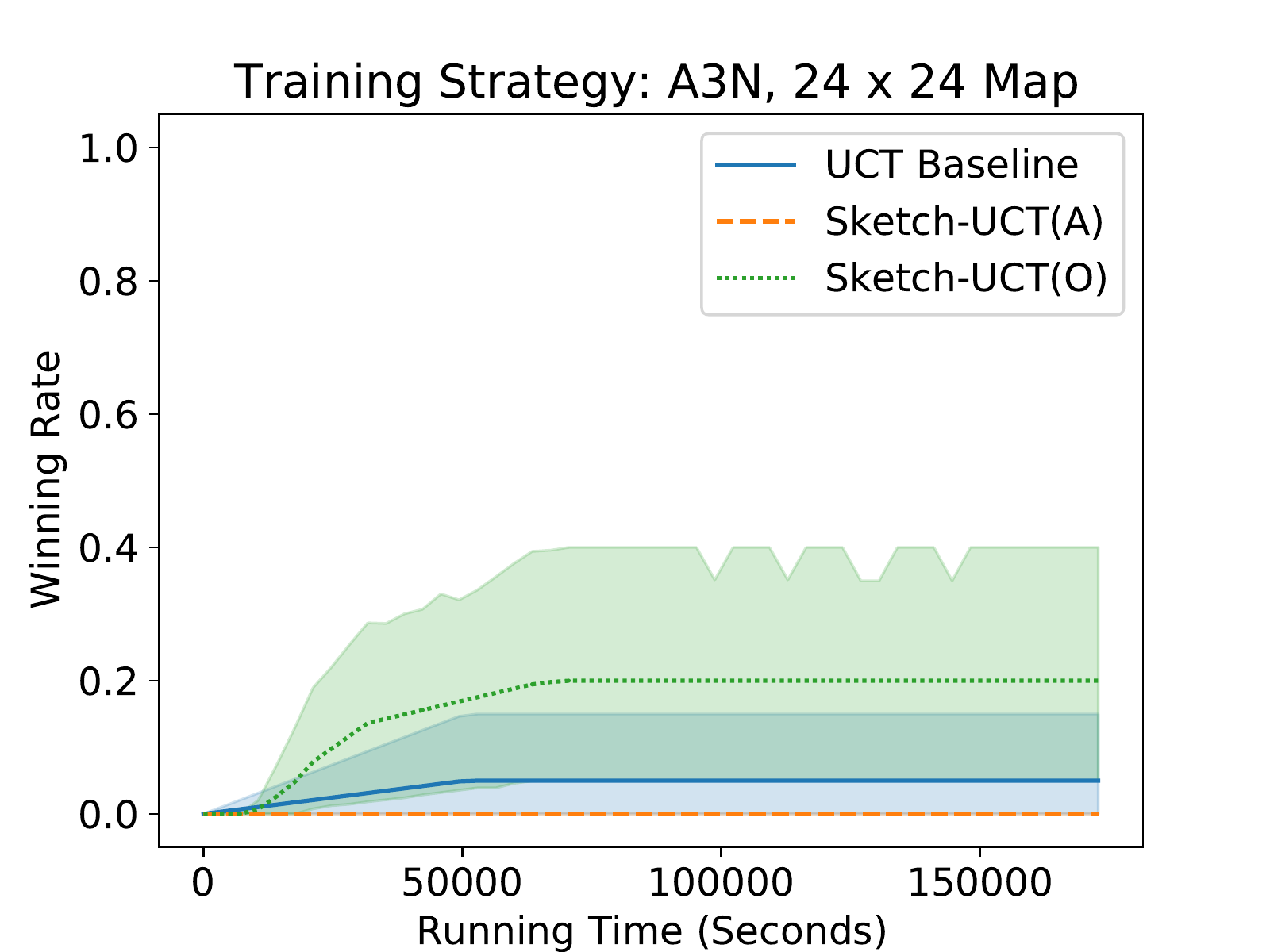}
    \includegraphics[width=.24\textwidth]{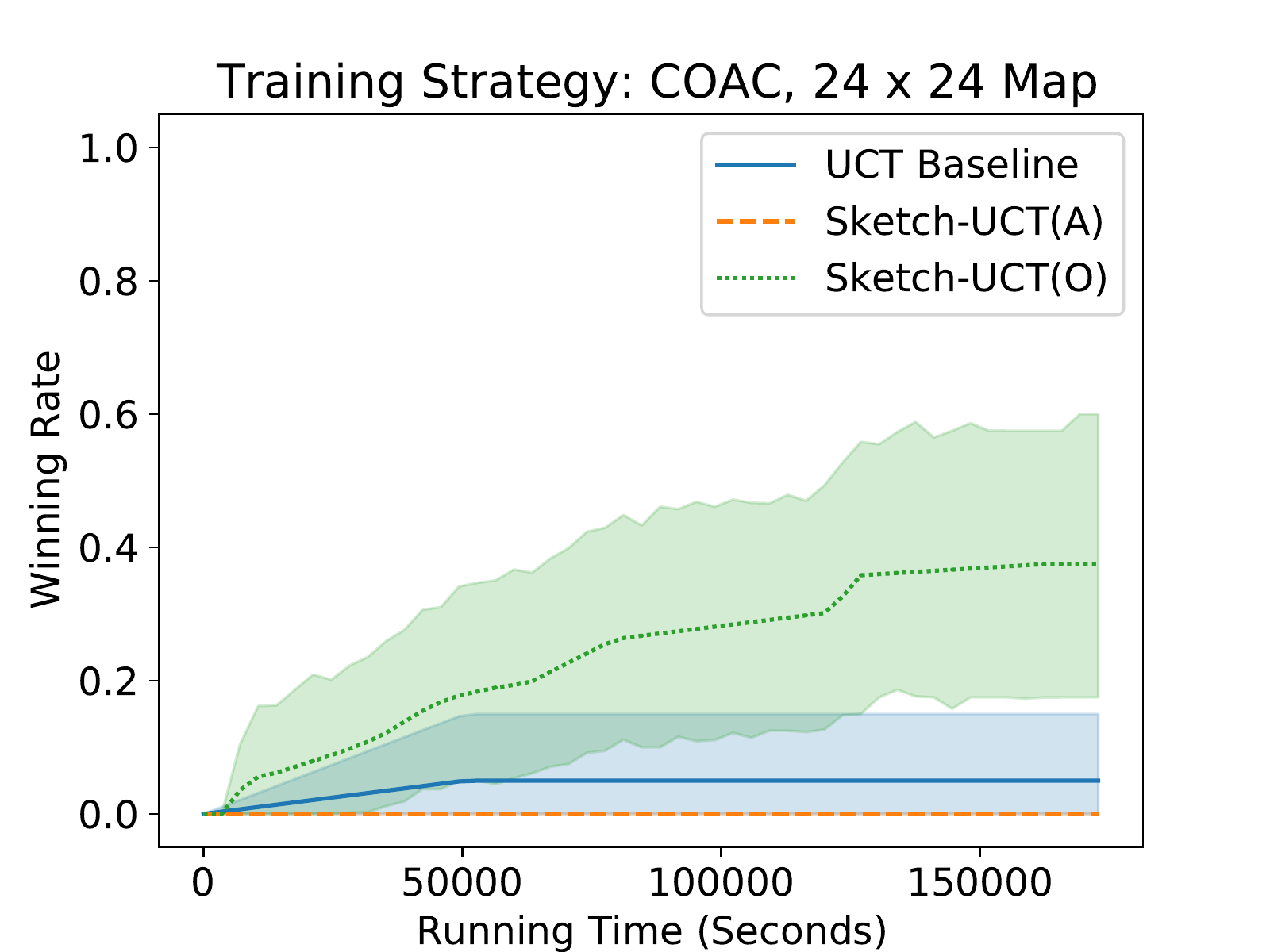}
    \caption{Winning rate of strategies UCT variants synthesize; the maps increase in size from left to right and top to bottom.}
    \label{fig:micrortsUCT}
\end{figure*}

Figure~\ref{fig:micrortsUCT} shows the results of the UCT variants on MicroRTS. Like in Can't Stop and the SA variants on MicroRTS, the sketch-based methods are better than the baseline in the 16x16 and 24x24 maps. None of the methods, including the UCT Baseline, were able to synthesize programs that defeat $\sigma_{-i}$ in the 32x32 and 64x64 maps given the time limit specified, thus we omit their plots. 


\section{Winning Rate and Cloning Score Correlation}

\begin{figure}[t]
\begin{center}
\includegraphics[width=230px]{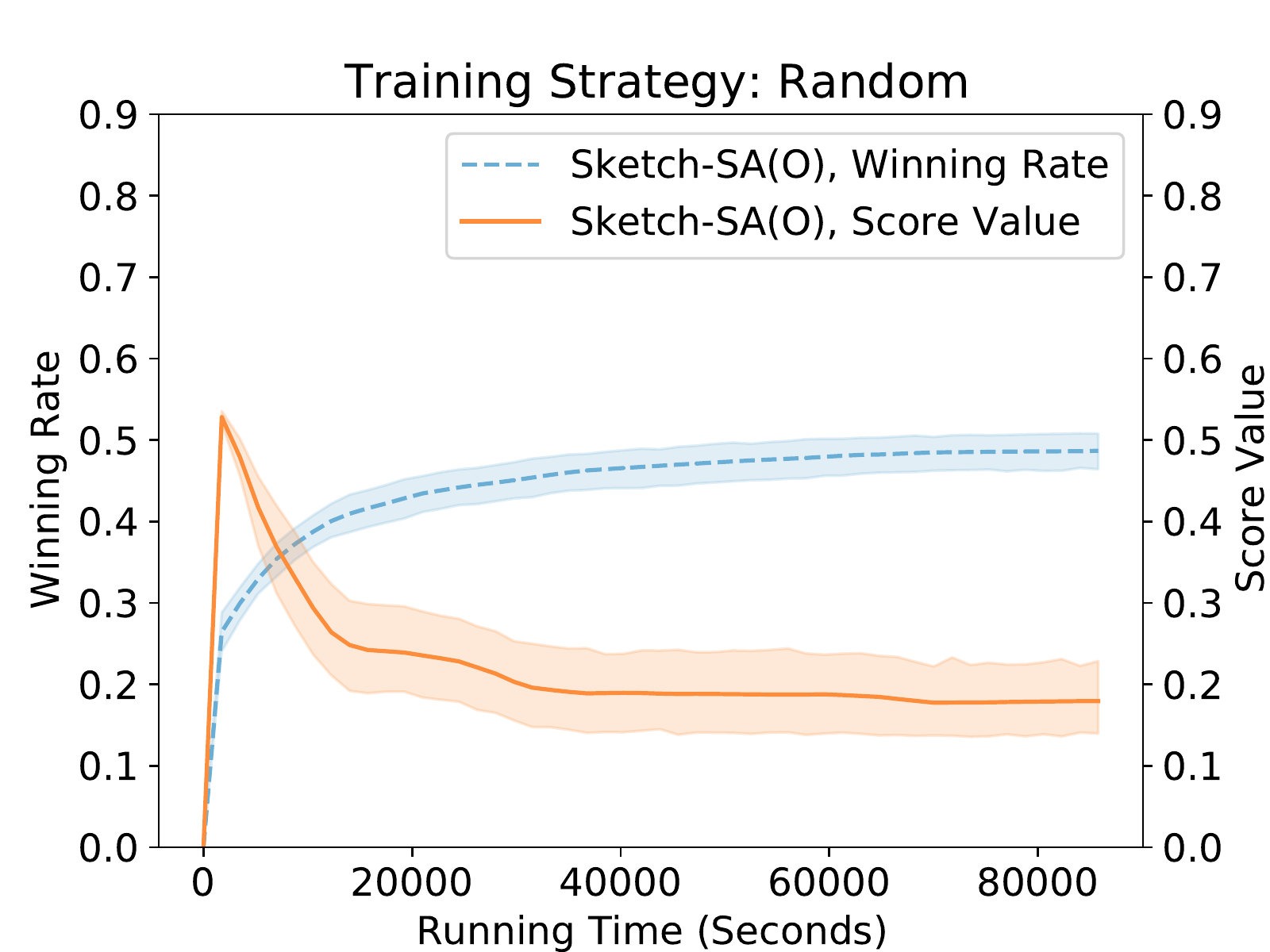}
\end{center}
\caption{Winning rate and cloning score value of the best program encountered during synthesis.}
\label{fig:correlation_state_random}
\end{figure}

\begin{figure}[t]
\begin{center}
\includegraphics[width=210px]{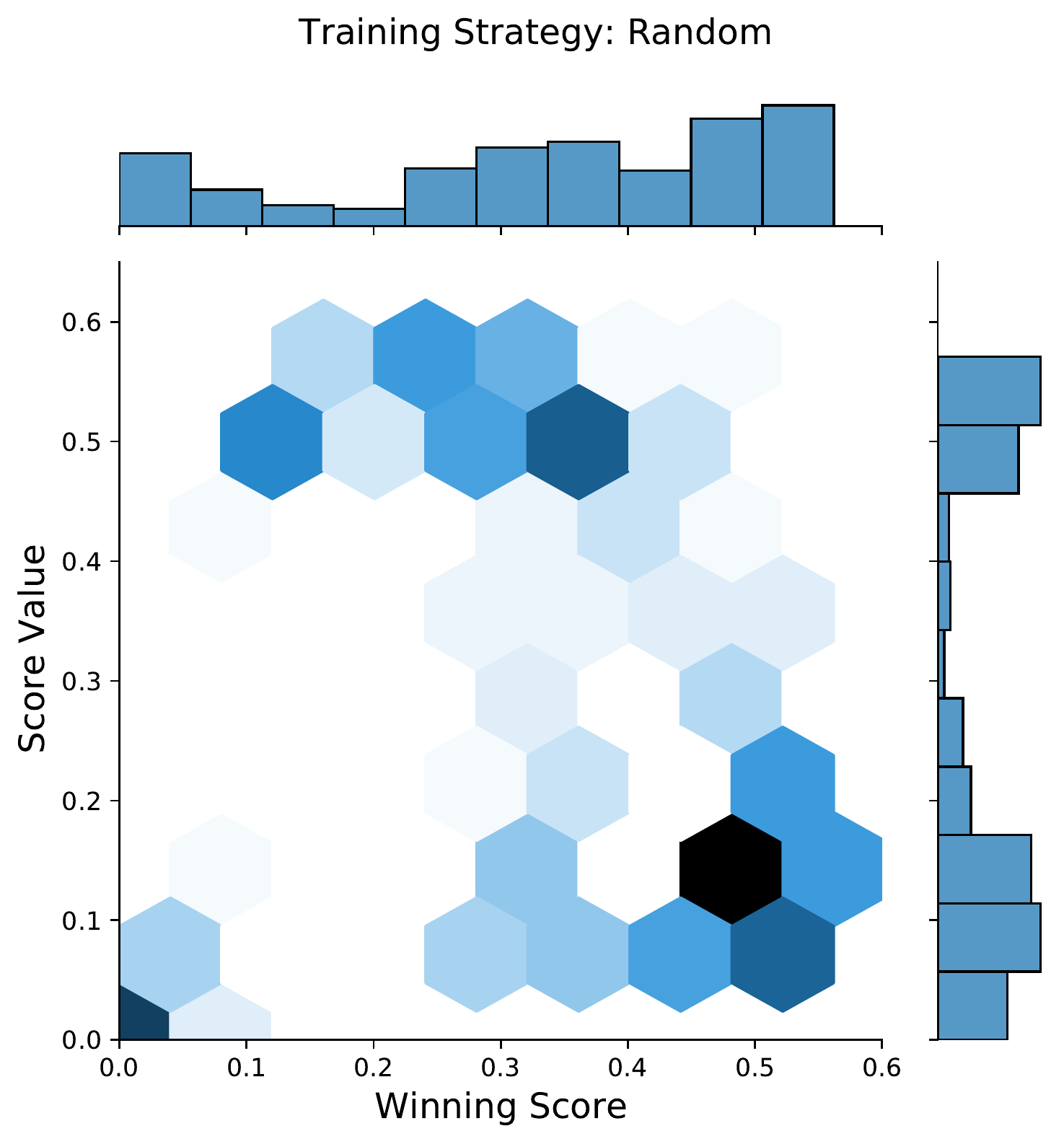}
\end{center}
\caption{Distribution of programs according to their winning score and cloning score.}
\label{fig:heatmap_state_random}
\end{figure}

 
 Figure~\ref{fig:correlation_state_random} shows the relation between the winning rate (left y-axis) and the observation cloning score (right y-axis) for the best programmatic strategy encountered during the SA searches (sketch-search and BR-search) for the game of Can't Stop. The cloned strategy is Random. The spike in the score value in the beginning of the search represents the end of sketch-search, which optimizes for the cloning score. The BR-search starts where the score value drops. Initially, the winning rate also grows with the score value, as the synthesizer is learning the basic structure of the final program. Once we start the BR-search the two curves are no longer correlated (the score decreases while winning rate increases). This is expected, as the strategy would become more and more similar to the Random strategy, which is a weak strategy, if the score value continued to grow. 
 
 

 Figure~\ref{fig:heatmap_state_random} presents a similar story by showing the distribution of a set of programmatic strategies according to their cloning score (y-axis) and winning rate against Glenn and Aloi's strategy (x-axis). The strategies with higher winning rates are not similar to the Random strategy (see the darker colors at the bottom-right corner). Despite the lack of correlation between winning rate and cloning score, cloning the behavior of Random can still help the search by allowing it to find helpful program sketches. 

\section{Sketch-Search-Only Results}

Table \ref{tab:sketchsearchonly} shows the results of solely using the behavioral cloning evaluation function $C(L,p)$ on the Can't Stop to guide the SA search. The table shows the winning rate and standard deviation against Glenn and Aloi's strategy of the strategy the system returns on 30 independent runs.

\begin{table*}[ht]
\sisetup{separate-uncertainty}
\centering
\begin{tabular}
  {@{} 
    P{3.5cm}
    S[table-format = 14]
    S[table-format = 2.3(3)]
    S[table-format = 2.3(3)]
  @{}}
  \toprule
   {Observation Type} & {$L$} & {BC} & {Sketch-SA} \\
  \midrule
    \multirow{3}{*}{A} & {Human} & 0.352(199) & 0.491(117)  \\
     &  {Glenn and Aloi} &  0.424(070) &  0.516(092)  \\
     &  {Random} & 0.182(058) & 0.507(085)\\
  \midrule
    \multirow{3}{*}{O} & {Human} & 0.212(216) & 0.555(017)  \\
     &  {Glenn and Aloi} &  0.368(088) &  0.543(024)  \\
     &  {Random} & 0.244(114) & 0.518(058) \\
  \bottomrule
\end{tabular}
\caption{\label{tab:sketchsearchonly} Winning rate against GA of strategies synthesized with an approach that uses only behavioral cloning (BC) and strategies synthesized with our proposed sketch-based methods (Sketch-SA) using different data sets $L$.}
\end{table*}


SA guided solely by the $C$ function synthesized stronger strategies while cloning the Glenn and Aloi's strategy. 
We conjecture that because the DSL we used allows Glenn and Aloi's strategy to be synthesized exactly, it can then be able to synthesize programs that accurately clone the target strategy if given enough time and data, as opposed to the human's strategy, which is unlikely to be representable in our DSL.

\section{Programs Synthesized for Can't Stop}

We use the structure of \citeauthor{Glenn2009AGH}~\shortcite{Glenn2009AGH}'s strategy and the synthesizers need to generate a program that fill the holes responsible for the yes-no and column decisions. The program with holes is shown below, with the two question marks denoting the instructions that need to be generated. 
\begin{lstlisting}
def  get_action(self, state):
    actions = state.available_moves()
    if actions == ['y', 'n']:
        score =  ?
        if win_after_n(state):
            return 'n'
        elif available_columns(state):
            return 'y'
        else:
            if score >= 29:
                return 'n'
            else:
                return 'y'
    else:
        index = ?
        return actions[index]
\end{lstlisting}

The method \texttt{win\_after\_n(state)} checks if the player  will win if they choose to end their turn and the method  \texttt{available\_columns(state)} checks if there are available columns for the player to choose given the current board and dice configuration. 
%
%
The pseudocode below presents Glenn and Aloi's strategy written in our DSL. 

\begin{lstlisting}
def  get_action(self, state):
    actions = state.available_moves()
    if actions == ['y', 'n']:
        score =  sum(map(lambda x: ($f_1$+1)*$l_4$, $l_2$)) + $f_5$
        if win_after_n(state):
            return 'n'
        elif available_columns(state):
            return 'y'
        else:
            if score >= 29:
                return 'n'
            else:
                return 'y'
    else:
        index = argmax(map(lambda x: sum(map(lambda x: $f_2$ * $l_5$ - 6 * $f_6$, $l_1$)), $l_3$))
        return actions[index]
\end{lstlisting}



The following program was synthesized by Sketch-SA(O) and achieved the winning rate of $59.64\%$ against Glenn and Aloi's strategy in $5,000$ matches of the game. 


\begin{lstlisting}
yes-no: (($f_5$ * $f_5$) - (sum(map((lambda x : sum($l_2$)), $l_2$)) - ($f_5$ + sum(map((lambda x : ($l_4$ * ($f_1$ * $f_3$))), $l_2$)))))
column: argmax(map((lambda x : sum(map((lambda x : ($f_3$ + $l_5$)), $l_1$))), $l_3$))
\end{lstlisting}
For the yes-no decision, the synthesized program has a similar structure to Glenn and Aloi's strategy. Namely, it accounts for the columns in which the neutral tokens are positioned and the position of the neutral tokens in each column. This strategy differs from Glenn and Aloi's because it places a higher weight on the difficulty of a state when deciding whether to continue or to stop playing ($f_5$ appears as a quadratic term in the synthesized program, while it is linear in Glenn and Aloi's strategy). The synthesized strategy accounts for the number of rows conquered in previous rounds ($f_3$ is used in the synthesized strategy, but not in Glenn and Aloi's). 


The synthesized strategy for the column decisions fixes a weakness in Glenn and Aloi's strategy. Namely, their strategy ignores the rows the player has conquered in previous rounds of the game. By contrast, the synthesized strategy uses function $f_3$ in the \texttt{argmax} operator. While both strategies prefer even-numbered columns, in some games, depending on the dice roll, the player might have to play on odd-numbered columns. Glenn and Aloi's strategy continues to prefer even-numbered columns, even if they have achieved a promising position on odd-numbered ones. The synthesized strategy eventually prefers to continue playing on odd-numbered columns if they have conquered a ``sufficiently large'' number of rows. 



\section{Programs Synthesized for MicroRTS}

We present below the programs Sketch-SA(O) synthesized for the maps used in our experiments. In contrast with the main text, the programs were not simplified to improve readability. 

\begin{lstlisting}
def Sketch-SA-O-16x16(state s):
    for u in s:
        u.train(Ranged, Right, 4)
        u.harvest(9)
        u.attack(Closest)
        u.train(Worker, Up, 1)
        
def Sketch-SA-O-24x24(state s):
    for u in s:
        if not u.isBuilder():
       	    u.moveToUnit(Ally, LessHealthy)
        u.train(Ranged,Left,15)
        for u in s:
       	    u.idle()
        u.build(Barracks, EnemyDir, 100)
    for u in s:
        u.harvest(4)
        u.attack(LessHealthy)


def Sketch-SA-O-32x32(state s):
    for u in s:
        for u in s:
       	    for u in s:
       		    u.harvest(2)
       	    for u in s:
       		    u.train(Ranged,EnemyDir,6)
        u.train(Worker,Left,7)
        for u in s:
       	    u.build(Barracks,Right,1)
        u.idle()
        u.harvest(5)
        u.attack(Strongest)

def Sketch-SA-O-64x64(state s):
    for u in s:
        u.attack(Weakest)
        for u in s:
           	u.train(Worker, Right, 6)
           	u.harvest(7)
           	for u in s:
           		u.build(Barracks, Left, 10)
           	if HaveUnitsAttacking(4):
           		 u.attack(Farthest)
           	else:
           		u.idle()
           	u.train(Ranged, Down, 50)
\end{lstlisting}

\section{MicroRTS Maps}

\begin{figure}[t]
    \centering
    \includegraphics[width=.33\textwidth]{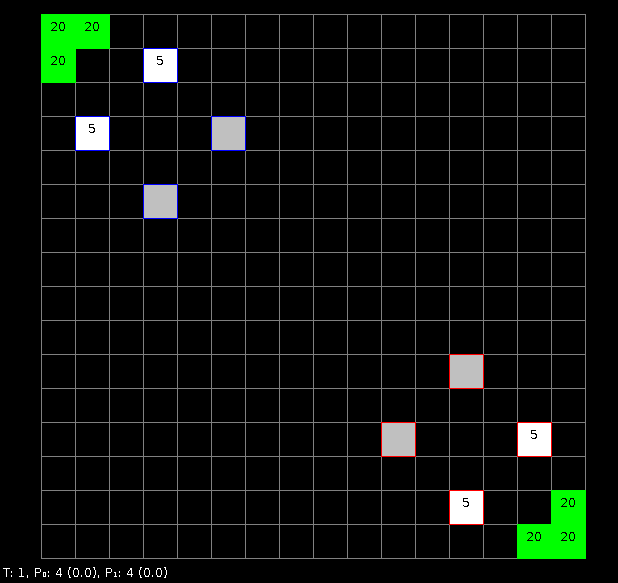}
    \includegraphics[width=.33\textwidth]{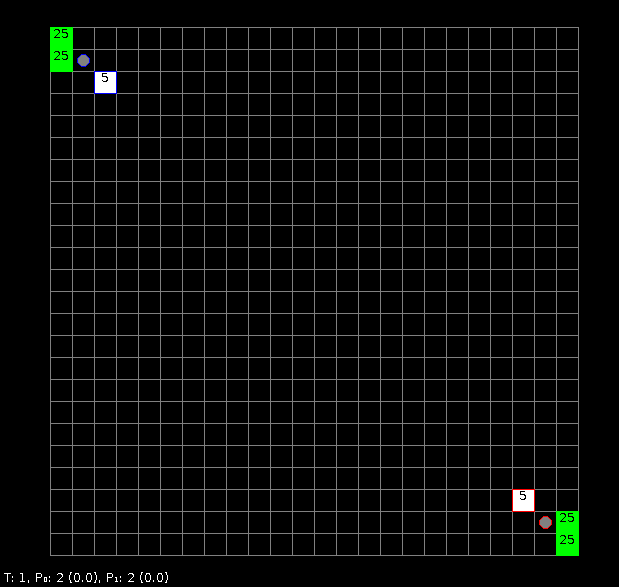}
    \includegraphics[width=.33\textwidth]{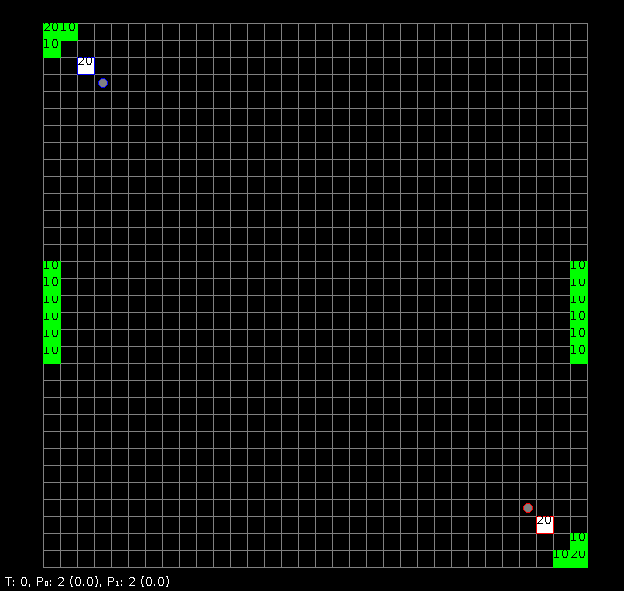}
    \includegraphics[width=.33\textwidth]{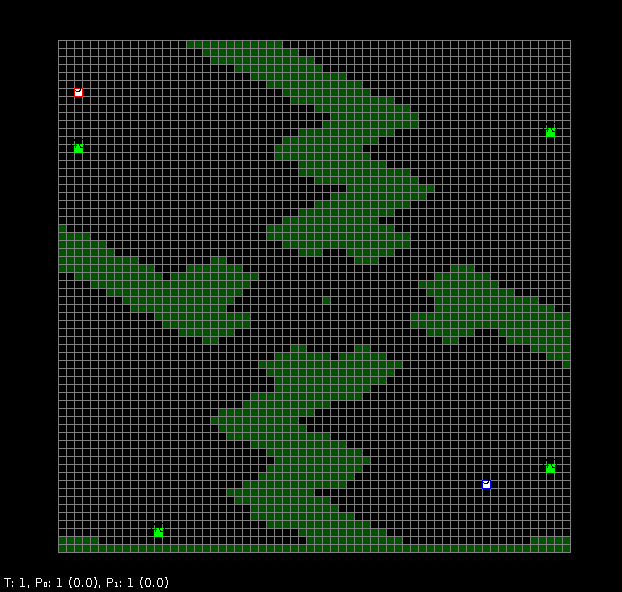}
    \caption{Maps used in the experiments. From top to bottom: 16$\times$16, 24$\times$24, 32$\times$32 and 64$\times$64.}
    \label{fig:microrts_maps}
\end{figure}

Figure~\ref{fig:microrts_maps} shows the MicroRTS maps used in our experiments. The green squares indicate resources that Workers can harvest. The white squares represent Bases that are used to store resources and train Workers. The gray squares are Barracks that are used to train attacking units. 

\end{document}